\pdfoutput=1

\documentclass[11pt]{article}

\usepackage{ACL2023}

\usepackage{times}
\usepackage{latexsym}

\usepackage{amsmath}
\usepackage{amssymb}
\usepackage{mathtools}
\usepackage{amsthm}

\usepackage[T1]{fontenc}

\usepackage[utf8]{inputenc}

\usepackage{microtype}

\usepackage{titlesec}
\titlespacing*{\subsubsection}{0pt}{1.0ex plus 0.2ex minus .2ex}{0.5ex plus .2ex}

\usepackage{inconsolata}
\usepackage{booktabs}
\usepackage{tabularx}
\usepackage{paralist}
\usepackage{multirow}
\usepackage{colortbl}
\usepackage{graphicx}
\usepackage[normalem]{ulem}

\usepackage{tcolorbox}
\usepackage{adjustbox}

\usepackage{enumitem}
\usepackage{pgfplots}
\pgfplotsset{compat=1.17}
\tcbuselibrary{breakable}

\definecolor{nmgray}{RGB}{229,229,229}

\tcbset{
    colback=gray!5,
    colframe=black!75!black,
    before skip=2pt, 
    after skip=2pt, 
    left=0pt,
    right=0pt,
    top=0pt,
    bottom=0pt,
    boxsep=5pt, 
    arc=0pt, 
    boxrule=1pt, 
}

\newtcolorbox{mybox}[2][]{
width=\columnwidth,
colback = nmgray!75!white, 
colframe = nmgray!75!white, 
boxsep=0pt,left=9pt,right=10pt,top=0pt,bottom=0pt,
fontupper=\linespread{0.9}\selectfont,
title=#2,#1}

\title{Faithful Logical Reasoning via Symbolic Chain-of-Thought}

\author{
Jundong Xu\textsuperscript{\rm 1},  \,
Hao Fei\textsuperscript{\rm 1}\Thanks{ Corresponding author},  \,
Liangming Pan\textsuperscript{\rm 2},  \,
Qian Liu\textsuperscript{\rm 3},  \,
Mong-Li Lee\textsuperscript{\rm 1},  \,
Wynne Hsu\textsuperscript{\rm 1} \\
\textsuperscript{\rm 1} National University of Singapore, Singapore \\
\textsuperscript{\rm 2} University of California, Santa Barbara, USA \\
\textsuperscript{\rm 3} University of Auckland, New Zealand \\
\tt {jundong.xu@u.nus.edu; haofei37@nus.edu.sg; liangmingpan@ucsb.edu} \\
\tt {liu.qian@auckland.ac.nz; dcsleeml@nus.edu.sg; whsu@comp.nus.edu.sg}
}

\begin{document}
\maketitle

\begin{abstract}
While the recent Chain-of-Thought (CoT) technique enhances the reasoning ability of large language models (LLMs) with the theory of mind, it might still struggle in handling logical reasoning that relies much on symbolic expressions and rigid deducing rules.
To strengthen the logical reasoning capability of LLMs, we propose a novel Symbolic Chain-of-Thought, namely \textbf{SymbCoT}, a fully LLM-based framework that integrates symbolic expressions and logic rules with CoT prompting. 
Technically, building upon an LLM, SymbCoT 1) first translates the natural language context into the symbolic format, and then 2) derives a step-by-step plan to solve the problem with symbolic logical rules, 3) followed by a verifier to check the translation and reasoning chain.
Via thorough evaluations on 5 standard datasets with both First-Order Logic and Constraint Optimization symbolic expressions, SymbCoT shows striking improvements over the CoT method consistently, meanwhile refreshing the current state-of-the-art performances.
We further demonstrate that our system advances in more faithful, flexible, and explainable logical reasoning.
To our knowledge, this is the first to combine symbolic expressions and rules into CoT for logical reasoning with LLMs.
Code is open at \url{https://github.com/Aiden0526/SymbCoT}.
\end{abstract}

\begin{figure}[!t]
\centering
\includegraphics[width=0.98\columnwidth]{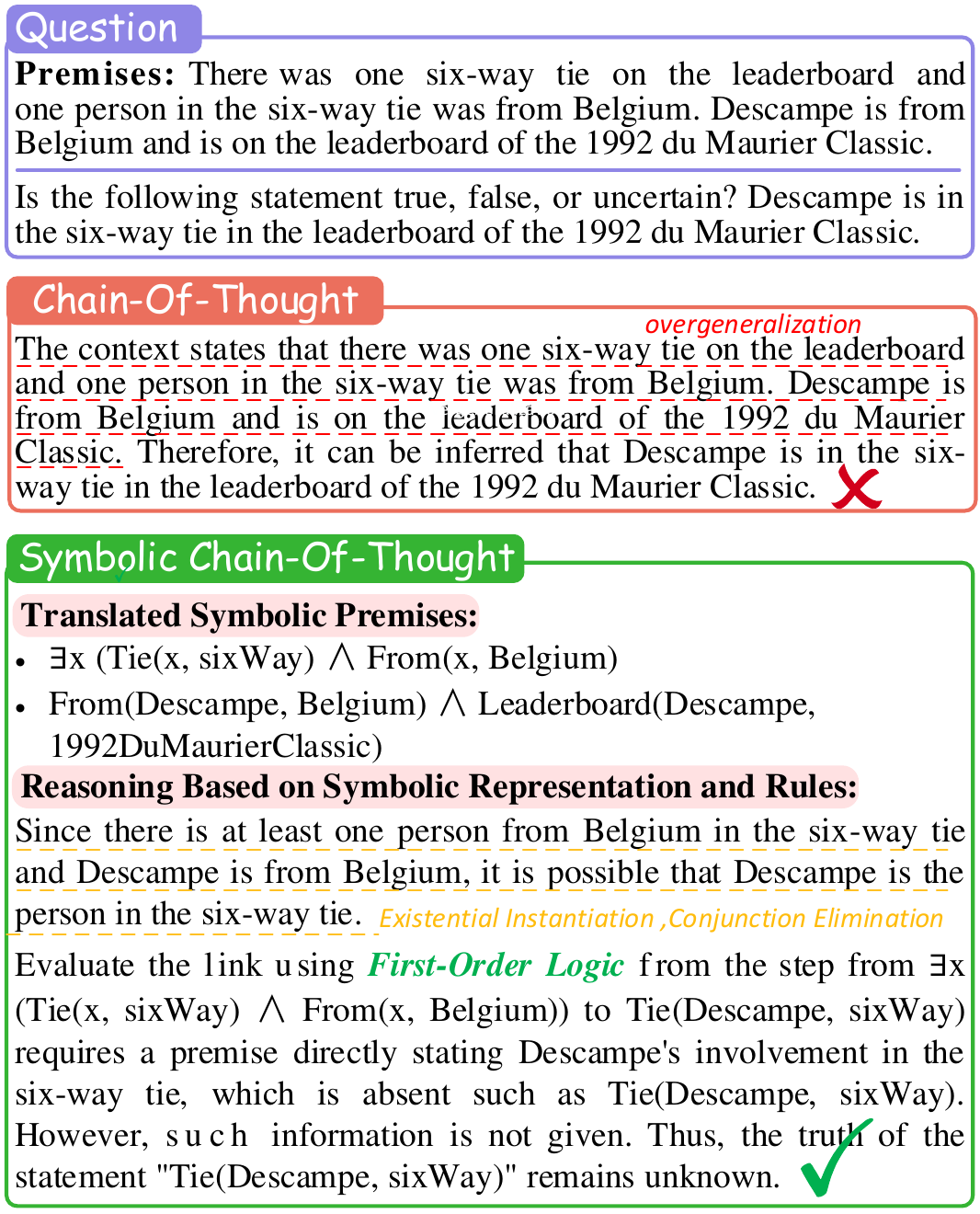}
\vspace{-2mm}
\caption{
An illustrative example of logical reasoning via Chain-of-Thought and our proposed Symbolic CoT (SymbCoT).
}
\label{fig:intro}
\vspace{-4mm}
\end{figure}

\section{Introduction}

Achieving human-like logical reasoning capabilities is crucial for realizing AGI, which plays a pivotal role in enabling intelligent systems to engage in problem-solving, decision-making, and critical thinking.
Recently, LLMs \cite{gpt3, palm} have demonstrated unprecedented capabilities in semantic understanding, casting a beacon of hope toward achieving AGI. 
Further enhancing LLMs to achieve human-level reasoning abilities, particularly in logical reasoning, is of paramount importance. 
Logical reasoning \cite{logic-reasoning1} stands out as a quintessential form of reasoning that, unlike other types, is crucial and challenging. 
It epitomizes a cognitive process characterized by rigorous evidence evaluation, argument construction, and logical deduction \cite{logic-reasoning2}. 
The latest trend is integrating LLMs with symbolic solvers to enhance their performance \cite{linc,pan-etal-2023-logic}. 
Unfortunately, these efforts have been limited to using LLMs merely as text-to-symbolic translators, with the core reasoning still reliant on traditional external reasoners \cite{prover9}. 
Such an approach, first, does not intrinsically strengthen LLMs' capability in logical reasoning.
Besides, over-reliance on external symbolic solvers often results in inflexibility, information omission, and unexplainability.

\begin{figure*}[!t]
\centering
\includegraphics[width=0.98\textwidth]{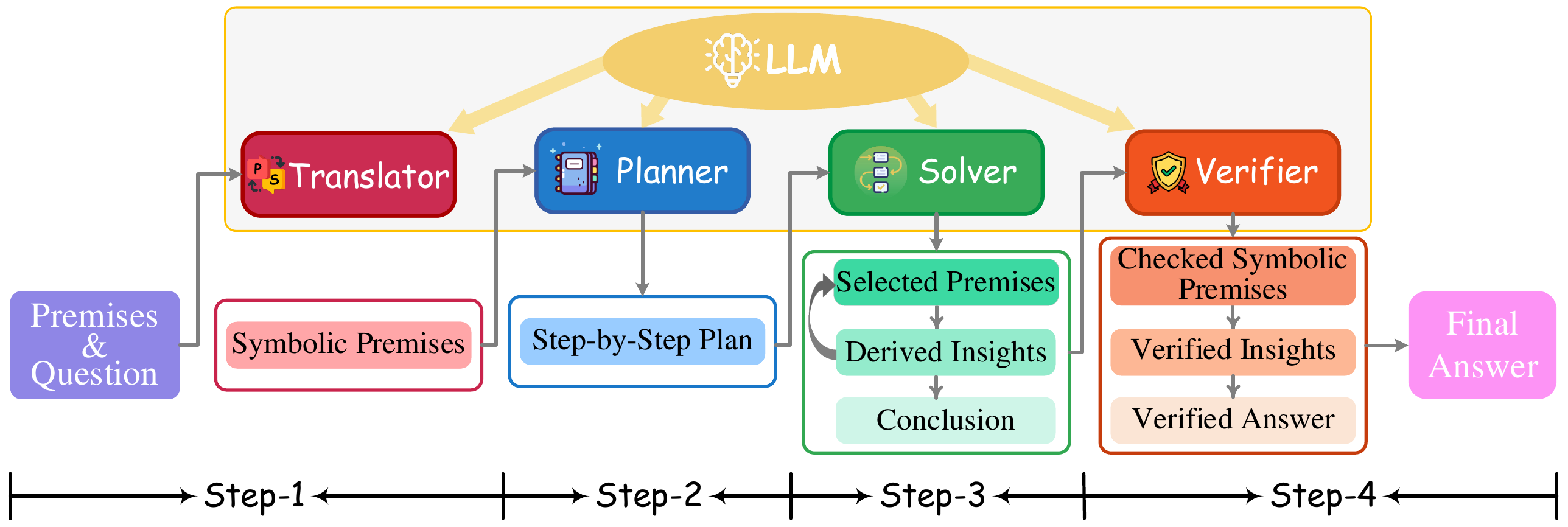}
\vspace{-2mm}
\caption{
Overview of the workflow in our proposed symbolic CoT framework. 
}
\label{fig:framework}
\vspace{-4mm}
\end{figure*}

On another note, the concept of CoT \cite{cot} has been introduced to mimic human thinking processes by encouraging LLMs to explicitly consider intermediate steps during problem-solving and to provide rationales for decisions, thereby enhancing the reliability of the reasoning process.  
CoT has been successfully integrated into a wide array of tasks \cite{auto-cot,implicit_cot, multitool-cot}, significantly improving LLMs' reasoning capabilities, sometimes even matching human performance in certain scenarios \cite{human-performance}. 
There is growing interest in applying CoT for logical reasoning \cite{logic-reasoning}, and developing advanced strategies such as self-consistency \cite{CoT-SC} and Tree-of-Thought \cite{ToT} for enhancement. 
However, applying basic CoT directly to logical reasoning is inherently limited, due to the abstractive nature of language expression.
Logical reasoning demands rigorous logical calculations, heavily relying on both symbolic expressions and rigid deducing rules to represent the internal structure of problems. 
Plain texts often fall short of supporting such precise logic, especially in scenarios that demand strict logical representation.
For instance, as shown in Fig. \ref{fig:intro}, when tackling a logical reasoning problem, utilizing symbolic representations like First-Order Logic (FOL) is more representative and precise than fully natural language rationales in CoT, enabling strict logical reasoning through clear inference rules.

To address these challenges, we introduce a novel Symbolic CoT (namely \textbf{SymbCoT}) for logical reasoning. 
Unlike existing state-of-the-art (SoTA) LLM-based symbolic reasoning systems \cite{linc,pan-etal-2023-logic}, SymbCoT is entirely facilitated by LLMs without relying on any external reasoners/tools, i.e., encompassing both the initial translation and subsequent reasoning phases. 
Fig. \ref{fig:framework} provides a high-level illustration of the overall system workflow. 
Technically, SymbCoT comprises four main modules: \emph{Translator}, \emph{Planner}, \emph{Solver}, and \emph{Verifier}. 
Notably, SymbCoT is characterized by the following three core aspects:
\vspace{-2.5mm}
\setdefaultleftmargin{1.5em}{2.2em}{1.87em}{1.7em}{1em}{1em}
\begin{compactitem}
    \item[1)] SymbCoT integrates symbolic expressions into CoT to describe intermediate reasoning processes, facilitating more precise logical calculations. However, relying solely on symbolic representation still has its limitations, as it often fails to capture certain content, such as implicit intentions or crucial contextual information embedded within questions. 
    Yet LLMs excel at interpreting such nuanced information and contexts. 
    Thus, we consider a combination of symbolic and natural language expressions to leverage the mutual strengths of both: freely expressed implicit intents and contextual information in natural language and rigorous expression in symbolic forms.

    \item[2)] Unlike the straightforward prompting of ``\emph{thinking step by step}'' in vanilla CoT, SymbCoT considers a \emph{plan-then-solve} architecture. 
    This involves decomposing the original complex problem into a series of smaller, more manageable sub-problems, which are then addressed one by one. 
    This way, the entire reasoning process becomes more trackable, enabling a clearer and more structured approach to problem-solving.

    \item[3)] Furthermore, we devise a retrospective verification mechanism. At both the translation and subsequent problem-solving stages, we retrospectively validate the correctness of each step's outcome, by tracing back to the original given condition. 
This verification process ensures the accuracy and reliability of the operations performed during the reasoning process.
        
\end{compactitem}

In experiments, we test SymbCoT with symbolic expressions of FOL and Constraint Optimization (CO) on five logical reasoning datasets using both GPT-3.5 and GPT-4. 
Results demonstrate that SymbCoT significantly enhances the reasoning capabilities of vanilla CoT, outperforming current SoTA solutions clearly. 
We further demonstrate that the more complex the logical reasoning task, the more pronounced the improvement of SymbCoT over vanilla CoT, further with the verification mechanism ensuring the faithfulness of the reasoning process. 
Our in-depth analysis reveals that fully LLM-based logical reasoning can offer better symbolic syntax robustness, human-readable explanations, and fuller utilization of information.

In summary, our technical contributions are: 
\begin{compactitem}

    \item proposing a fully LLM-based logical reasoning framework based on CoT, demonstrating that LLMs can achieve robust logical reasoning capabilities without external reasoning tools. 
    Compared to existing SoTA solutions relying on external resolvers,
    SymbCoT offers better robustness against translation errors and more human-understandable explanations.

    \item innovatively integrating the strengths of symbolic forms and natural language expressions, enabling precise reasoning calculations while fully interpreting implicit information and capturing rich contexts. 

    \item introducing a plan-then-solve architecture for CoT reasoning, along with a retrospective verification mechanism, enhancing the faithfulness of the reasoning process.

\end{compactitem}

\section{Related work}

\vspace{-2mm}
Recent achievements in reasoning research powered by LLMs have shown promising results \cite{logic-reasoning,human-performance}, bringing LLMs closer to human-level reasoning capabilities due to their profound semantic understanding \cite{wu2023next,FeiMatchStruICML22}.
Among these, the CoT series methodology \cite{cot} has garnered increasing attention for its emulation of human discrete chain reasoning. 
By considering more intermediate steps and the rationales behind decision-making, CoT has significantly enhanced overall reasoning performance on many downstream applications \cite{implicit_cot,fei2024video}. 
Subsequent technologies have introduced more advanced reasoning frameworks, incorporating mechanisms such as self-consistency and non-linear, multidimensional topological structures, e.g., Tree-of-Thought \cite{ToT}, Graph-of-Thought \cite{got,zheng2024reverse}, and other variants \cite{auto-cot, plan-and-solve}.

However, research has also highlighted limitations within CoT due to its reliance on natural language rationales, which may not always be advantageous in certain scenarios. 
Studies have found that representing CoT's intermediate steps in a structured manner, reflecting the task's intrinsic structure, can bolster reasoning capabilities for specific tasks \cite{code-prompting,mathprompter}. 
For instance, using pseudo-code to describe intermediate reasoning processes has been shown to enhance outcomes in code generation tasks \cite{SCoT}, while adopting mathematical equations for CoT's steps has proven beneficial in solving mathematical problems \cite{mathprompter}.
Focusing on logical reasoning, it becomes evident that solely using natural language formats for intermediate reasoning steps inevitably leads to significant information loss, especially when tackling complex logical reasoning jobs. 
This paper, therefore, proposes a symbolic-oriented CoT approach tailored for logical reasoning.

Logical reasoning \cite{logic-reasoning1}, a paramount aspect of the reasoning domain, demands models that can precisely grasp and manipulate complex logical structures. 
Previous works have explored rule-based \cite{prover9} and neural-based solving \cite{neural-logic2,neural-logic} methods for interpreting symbolic representations. 
The latest trend involves integrating LLMs into the symbolic reasoning process \cite{llm_symbolic_generation, llm_symbolic_math}. 
For example, Logic-LM \cite{pan-etal-2023-logic} and LINC \cite{linc} consider using LLMs as translators to convert natural language into symbolic syntax such as FOL, which is then processed by external reasoning tools to enhance reasoning performance. 
These approaches maintain that LLMs cannot parse symbolic expressions as reliably as external rule-based reasoners.

Nonetheless, merely utilizing LLMs as translators does not inherently enhance their logical reasoning capabilities. 
This work pioneers the development of the first symbolic CoT specifically designed for logical reasoning, fully utilizing LLMs.
In contrast to approaches like Logic-LM and LINC, our method demonstrates several advancements:
First, external reasoners require strict formatting, where any translation error by LLMs can lead to failure in parsing and reasoning. 
Our reasoning steps, facilitated by the LLM, exhibit greater robustness against syntax errors.
Second, the entire reasoning process is conducted by the LLM, providing rationales that ensure a more human-friendly explanation throughout.
Third, we propose a blend of symbolic forms and natural language expressions within the logical reasoning process, achieving precise reasoning calculations while fully interpreting implicit information inherent in natural language.
Finally, we introduce a plan-then-solve CoT reasoning architecture and a verification mechanism, ensuring the faithfulness of the reasoning process.

\section{SymbCoT for Symbolic Reasoning}

\vspace{-2mm}
\subsection{Task Definition}

The logical reasoning is defined as: formally,
given a set of premises \(P = \{p_1, p_2, \ldots, p_n\}\), where each \(p_i\) represents a logical statement, we aim to derive a conclusion regarding a given statement \(S\). 
The objective is to determine whether \(S\) is true (\(T\)), false (\(F\)), or unknown (\(U\)) based on the logical inferences drawn from the premises.

\begin{tcolorbox}[fontupper=\linespread{0.7}\selectfont,]
{\footnotesize
$\blacktriangleright$ \textbf{Example:}\\
$<$Premises$>$ ($P$)\\
A hawk never lands. Some birds are hawks. \\
$<$Statement$>$ ($S$)\\  
All birds land. \\
$<$Answer$>$\\ 
False.
}
\end{tcolorbox}

\subsection{Modules}

\vspace{-2mm}
As shown in Fig. \ref{fig:framework}, our SymbCoT system is fully supported by LLMs and comprises four distinct modules: \textbf{Translator}, \textbf{Planner}, \textbf{Solver}, and \textbf{Verifier}, whose roles are elaborated as follows. \footnote{We use few-shot demonstrations for each module.}

\vspace{-2mm}
\paragraph{Translator}
converts the premises and a question statement from natural language to a symbolic format. 
This process prepares the input in a way that aligns with the structured requirements of subsequent reasoning processes, ensuring that the reasoning problems are represented in a format conducive to logical analysis.

\vspace{-2mm}
\paragraph{Planner}
breaks down the raw problem into smaller sub-problems, which develop a detailed, step-by-step plan that connects the given premises to the question statement in both natural language and symbolic form. 
This blueprint guides the structured reasoning process by outlining a logical sequence of steps.

\vspace{-2mm}
\paragraph{Solver}
is tasked with deriving the answer through a sequential logical inference process given the premises and question statement, along with a meticulously crafted plan. 
The Solver module emphasizes the importance of basing logical deductions on symbolic inference rules. 
For instance, the LLM is explicitly instructed to apply logical deduction techniques adhering to FOL such as Modus Tollens \cite{modus}, e.g., If $\neg B$, and $(A \to B)$, then $\neg A$. 
This ensures that each step in the reasoning process is grounded in established logical frameworks, to draw accurate and reliable conclusions.

\vspace{-2mm}
\paragraph{Verifier}
serves two functions in our framework.
First, it validates the correctness of symbolic translations by prompting the LLM to ascertain their semantic equivalence to the corresponding natural language. 
If discrepancies are identified, the language model refines the symbolic translation based on the original natural language to ensure semantic consistency.
Second, it verifies the output of the Solver containing step-by-step logical reasoning with the final answer. 
During this verification process, the Verifier scrutinizes each logic deduction made, to ensure they strictly comply with the principles of formal logic. 
If invalid logic is found, the Verifier refines the reasoning steps according to the correct logic to arrive at a new answer.
These ensure both the fidelity and reliability of symbolic translations and logical reasoning soundness.

\vspace{-2mm}
\subsection{Reasoning Steps}
We now walk through the SymbCoT reasoning steps with detailed examples, each corresponding to the process exemplified in Fig. \ref{fig:framework}.

\vspace{-2mm}
\paragraph{Step 1: Translating natural language context into symbolic.}
Given the original set of premises \(P = \{p_1, p_2, \ldots, p_n\}\) and the question statement \(S\), the Translator will first interpret them into the corresponding symbolic format, denoted as \(P' = \{p'_1, p'_2, \ldots, p'_n\}\) and \(S'\). 
We concatenate the natural language and symbolic expression \(P\) and \(P'\) as \(P_c\), \(S\) and \(S'\) as \(S_c\), i.e., hybrid format expressions for the future steps of reasoning.

\begin{tcolorbox}[breakable, fontupper=\linespread{0.7}\selectfont,]
{\footnotesize
$\blacktriangleright$ \textbf{Input:}\\
Please parse the context and question into First-Order Logic formulas. \\
$<$Premises$>$ ($P$)\\ 
If a cartoon character is yellow, it is from the Simpsons. \\
If a cartoon character is from Simpsons, then it is loved by children. \\
(... More premises ...) \\
$<$Statement$>$ ($S$)\\
Ben is ugly or yellow. \\

$\blacktriangleright$ \textbf{Output:}\\
$<$Premises$>$ ($P'$)\\
$\forall x \, (Yellow(x) \rightarrow Simpsons(x))$ \\
$\forall x \, (Simpsons(x) \rightarrow Loved(x))$ \\
(... More premises ...) \\
$<$Statement$>$ ($S'$)\\
$(Yellow(ben) \lor Ugly(ben))$
}
\end{tcolorbox}

\paragraph{Step 2: Deriving plan based on the natural and symbolic context.}

At this stage, the Planner module, utilizing \(P_c\) and \(S_c\), constructs a plan comprising intermediate steps \(I = \{i_1, i_2, \ldots, i_m\}\). 
These steps form logical inferences sequentially bridging the premises \(P_c\) towards the question statement \(S_c\).

\begin{tcolorbox}[breakable, fontupper=\linespread{0.7}\selectfont,]
{\footnotesize
$\blacktriangleright$ \textbf{Input:}\\
Please derive a step-by-step plan using the First-Order Logic rule for determining the conclusion based on the context. \\

$<$Premises$>$ ($P_c$)\\ 
If a cartoon character is yellow, it is from the Simpsons ::: $\forall x \, (Yellow(x) \rightarrow Simpsons(x))$ \\
If a carton character is from Simpsons, then it is loved by children: $\forall x \, (Simpsons(x) \rightarrow Loved(x))$ \\
(... More premises ...) \\
$<$Statement$>$ ($S_c$)\\
Ben is ugly or yellow ::: $(Yellow(ben) \lor Ugly(ben))$ \\

$\blacktriangleright$ \textbf{Output:}\\
$<$Plan$>$ ($I$) \\
1: Identify the relevant premise of Ben. \\
2: Identify the relevant premise of yellow and ugly.\\
(... More plan ...)
}
\end{tcolorbox}

\paragraph{Step 3: Solving the problem given context and plan.}

Here, the Solver operates on the premises \(P_c\), the question statement \(S_c\), and the plan \(I\), by systematically applying logic rules. It iteratively selects pertinent premises and infers new insights through a sequence of reasoning steps, represented as \(D = \{d_1, d_2, \ldots, d_l\}\). Here, each \(d_k\) signifies an insight inferred from the relevant premises during the iteration process, contributing to the elucidation of \(S_c\). This iterative methodology leads to a conclusive determination \(C\), assigning the statement \(S_c\) as true (\(T\)), false (\(F\)), or uncertain (\(U\)).

\begin{tcolorbox}[breakable, fontupper=\linespread{0.7}\selectfont,]
{\footnotesize
$\blacktriangleright$ \textbf{Input:}\\
Please solve the question based on First-Order Logic rules such as Modus Ponens... \\
$<$Premises$>$ ($P_c$)\\ 
If a cartoon character is yellow, it is from the Simpsons: $\forall x \, (Yellow(x) \rightarrow Simpsons(x))$ \\
If a carton character is from Simpsons, then it is loved by children: $\forall x \, (Simpsons(x) \rightarrow Loved(x))$ \\
(... More premises ...) \\
$<$Statement$>$ ($S_c$)\\
Ben is ugly or yellow ::: $(Yellow(ben) \lor Ugly(ben))$ \\
$<$Plan$>$ ($I$)\\
1: Identify the relevant premise of Ben. \\
2: Identify the relevant premise of yellow and ugly.\\
(... More plan ...) \\

$\blacktriangleright$ \textbf{Output:}\\
$<$Reasoning Steps$>$ ($D$) \\ 
Step 1: Yellow(ben) $\rightarrow$ Simpsons(ben) by Modus Ponens from premise 1. \\
Step 2: Simpsons(ben) $\rightarrow$ Loved(ben) by Modus Ponens from premise 2. \\
(... More steps ...) \\
$<$Conclusion$>$ ($C$) \\
Thus, we can conclude that $(Yellow(ben) \lor Ugly(ben))$ is false by contradiction.
}
\end{tcolorbox}

\paragraph{Step4: Verifying the translation and solving process.}
Now based on the concatenated inputs \(P_c\) and \(S_c\) from Step 1, and the reasoning chain \(D\) and derived conclusion \(C\) from Step 3, the Verifier conducts two ways of verification process. 
First, it checks whether the symbolic language is correctly translated based on the original natural language context, represented as \(V_{\text{trans}} = \text{Verify}(P \rightarrow P', S \rightarrow S')\). 
If any incorrect translation is found, the Verifier will undertake refinements to produce a revised translation of the premises and statement, denoted as \(P'_{\text{refined}}\) and \(S'_{\text{refined}}\).
Second, it will check whether the inference \(D\) adheres to valid logical rules. 
If any logical fallacy is detected, the Verifier will conduct a refined derivation \(D_{\text{refined}}\) using the corrected logic rules.
Consequently, these refinements lead to the generation of a refined answer, represented as \(C_{\text{refined}}\).
Upon successful verification as \(V_{\text{validity}}(D/D_{\text{refined}})\), the Verifier produces a verified answer \(C_{\text{verified}}\).

\begin{tcolorbox}[breakable, fontupper=\linespread{0.7}\selectfont,]
{\footnotesize
$\blacktriangleright$ \textbf{Input:}\\
Please verify: 1) the symbolic context is consistent with the natural language; 2) the solving step is logically valid. \\
$<$Premises$>$ ($P_c$)\\ 
If a cartoon character is yellow, it is from the Simpsons: $\forall x \, (Yellow(x) \rightarrow Simpsons(x))$ \\
If a carton character is from Simpsons, then it is loved by children: $\forall x \, (Simpsons(x) \rightarrow Loved(x))$ \\
(... More premises ...) \\
$<$Statement$>$ ($S_c$)\\
Ben is ugly or yellow ::: $(Yellow(ben) \lor Ugly(ben))$ \\
$<$Reasoning Steps$>$ ($D$)\\
Step 1: Yellow(ben) $\rightarrow$ Simpsons(ben) by Modus Ponens from premise 1. \\
Step 2: Simpsons(ben) $\rightarrow$ Loved(ben) by Modus Ponens from premise 2. \\
(... More steps ...) \\
$<$Conclusion$>$ ($C$)\\
Thus, we can conclude that $(Yellow(ben) \lor Ugly(ben))$ is false by contradiction.\\

$\blacktriangleright$ \textbf{Output:}\\
$<$Translation Verification$>$ ($P/P'_{\text{refined}}$) and ($S/S'_{\text{refined}}$)\\
1. Verify context consistency: "If a cartoon character is yellow, it is from the Simpsons" is semantically consistent with "$\forall x \, (Yellow(x) \rightarrow Simpsons(x))$" \\
(... More translation verification ...) \\
$<$Logic Verification$>$ ($D/D_{\text{refined}}$)\\
2. Verify solving logic step by step: Step 1 correctly implement Modus Ponens using premise 1; Step 2 correctly implement Modus Ponens using premise 2; \\
(... More logic verification ...) \\
$<$Conclusion Verification$>$ ($C_{\text{verified}}$/$C_{\text{refined}}$) \\
Thus, the solving process is logically valid. The answer is verified to be false.
}
\end{tcolorbox}

\section{Experiments}

\vspace{-2mm}
\subsection{Settings}

\vspace{-1mm}
\paragraph{Model.} Following prior work of Logic-LM \cite{pan-etal-2023-logic}, we assess the baselines and our method using GPT-3.5 \cite{DBLP:conf/nips/Ouyang0JAWMZASR22} and GPT-4 \cite{gpt4}\footnote{Specifically, we use gpt-3.5-turbo-0613 and gpt-4-0613.}, ensuring a fair comparison with identical model parameters.

\vspace{-2mm}
\paragraph{Dataset.} Five standard datasets are employed, including PrOntoQA \cite{DBLP:conf/iclr/Saparov023}, ProofWriter \cite{proofwriter}, FOLIO \cite{folio}, LogicalDeduction \cite{DBLP:journals/corr/abs-2206-04615}, and AR-LSAT \cite{ar}. 
Each of them takes different symbolic representations and introduces its own set of challenges in the topic of logical reasoning.
The primary metric for evaluation is accuracy, measuring the multiple-choice correctness of the questions.

\vspace{-2mm}
\paragraph{Symbolic Structure.} In datasets PrOntoQA, ProofWriter, and FOLIO, we use FOL as symbolic structure.
To test the generalizability of our framework among different symbolic structures, we further consider the CO symbolic expression in datasets LogicalDeduction and AR-LSAT.

\begin{table}[!t]
  \centering
\fontsize{9}{9.5}\selectfont
\setlength{\tabcolsep}{1.5mm}
\begin{tabular}{lcccccccc}
\hline
 & ProntoQA & ProofWriter & FOLIO & Avg \\
\hline
\multicolumn{5}{l}{$\bullet$ \textbf{\emph{GPT-3.5-turbo}}} \\
Naive & 47.40 & 35.50 & 45.09 & 42.66 \\
CoT & \uline{67.80} & 49.17 & 57.35 & 58.11  \\
Logic-LM & 61.00 & \uline{58.33} & \bf 62.74 & \uline{60.69} \\
\rowcolor{nmgray}  \bf SymbCoT & \bf 75.80 & \bf 59.03 & \uline{57.84} & \bf 64.22  \\
\specialrule{0em}{-1pt}{-1pt} & \scriptsize{(+8.00)}  & \scriptsize{(+0.70)}  & \scriptsize{(-4.90)} & \scriptsize{(+3.53)} \\
\hline
\multicolumn{5}{l}{$\bullet$ \textbf{\emph{GPT-4}}} \\
Naive & 77.40 & 52.67 & 69.11 & 66.39 \\
CoT & \uline{98.79} & 68.11 & 70.58 & 79.16 \\
CoT-SC & - & 69.33 & 68.14 & - \\
ToT & - & 70.33 & 69.12 & - \\
CR & - & 71.67 & 69.11 & - \\
DetermLR & - & 79.17 & 75.45 & - \\
Logic-LM & 83.20 & \uline{79.66} & \uline{78.92} & \uline{80.59}  \\
\rowcolor{nmgray}  \bf SymbCoT & \bf 99.60 & \bf 82.50 & \bf 83.33 & \bf 88.47  \\
\specialrule{0em}{-1pt}{-1pt} & \scriptsize{(+0.81)}  & \scriptsize{(+2.84)}  & \scriptsize{(+4.41)} & \scriptsize{(+7.88)} \\
\hline
\end{tabular}%
\vspace{-2mm}
\caption{
Performance on symbolic reasoning with First-Order Logical representation. 
The second best score is \underline{underlined} and \textbf{bold} one is the best. In the brackets are the corresponding improvements in between.
}
\label{tab:main-1}
\vspace{-2mm}
\end{table}

\begin{table}[!t]
  \centering
\fontsize{9}{9.5}\selectfont
\setlength{\tabcolsep}{2mm}
\begin{tabular}{lcccccccc}
\hline
 & LogicalDeduction & AR-LSAT & Avg \\
\hline
Naive & 71.33 & 33.33 & 52.33 \\
CoT & 75.25 & 35.06 &  55.14 \\
CoT-SC & 74.67 & - & - \\
ToT & 76.83 & - & - \\
CR & 78.33 & - & - \\
DetermLR & 85.00 & - & - \\
Logic-LM & \uline{87.63} & \uline{43.04} & \uline{65.34} \\
\rowcolor{nmgray}  \bf SymbCoT & \bf 93.00 & \bf43.91 & \bf 68.46  \\
\specialrule{0em}{-1pt}{-1pt} & \scriptsize{(+5.37)}  & \scriptsize{(+0.87)}  & \scriptsize{(+3.12)} \\
\hline
\end{tabular}%
\vspace{-2mm}
\caption{
Results (using GPT-4) on symbolic reasoning with Constraint Optimization representation. 
}
\vspace{-4mm}
\label{tab:main-2}
\end{table}

\vspace{-2mm}
\paragraph{Baseline.} 
We compare with a range of established baselines.
Those based on GPT-3.5 are:
1) Naive Prompting;
2) CoT \cite{cot};
3) Logic-LM \cite{pan-etal-2023-logic}.
On GPT-4, apart from the above baselines, we further include more systems:
4) CoT-SC \cite{CoT-SC};
5) ToT \cite{ToT};
6) Cumulative Reasoning \citep[CR; ][]{CR};
7) DetermLR \cite{DetermLR};

Appendix extends more details of the symbolic structures ($\S$\ref{sec:symbolic_specific}), the datasets ($\S$\ref{dataset_specific}), and baselines ($\S$\ref{baseline_specific}), as well as the full prompt configurations ($\S$\ref{prompt_specific}).

\subsection{Main Result}
Table \ref{tab:main-1} shows that our method significantly outperforms Naive, CoT, and Logic-LM baselines, with gains of 21.56\%, 6.11\%, 3.53\% on GPT-3.5, and 22.08\%, 9.31\% and 7.88\% on GPT-4, respectively.
We notice the only exception is on the FOLIO dataset with GPT-3.5, failing to surpass Logic-LM.
The underperformance points to challenges in non-linear reasoning, reflecting the inherent challenge for LLMs.
But, our approach notably surpasses all baselines across both datasets with GPT-4, especially outperforming Logic-LM by an average of 7.88\%, which demonstrates significant improvements in complex reasoning tasks. 
In Table \ref{tab:main-2}, we show the results with CO symbolic expression on two datasets.
As seen, our approach surpasses both CoT and Logic-LM by 13.32\% and 3.12\%, respectively, again demonstrating its general versatility in different symbolic reasoning expressions.

\begin{figure}[!t]
\centering
\resizebox{0.98\columnwidth}{!}{
  \includegraphics{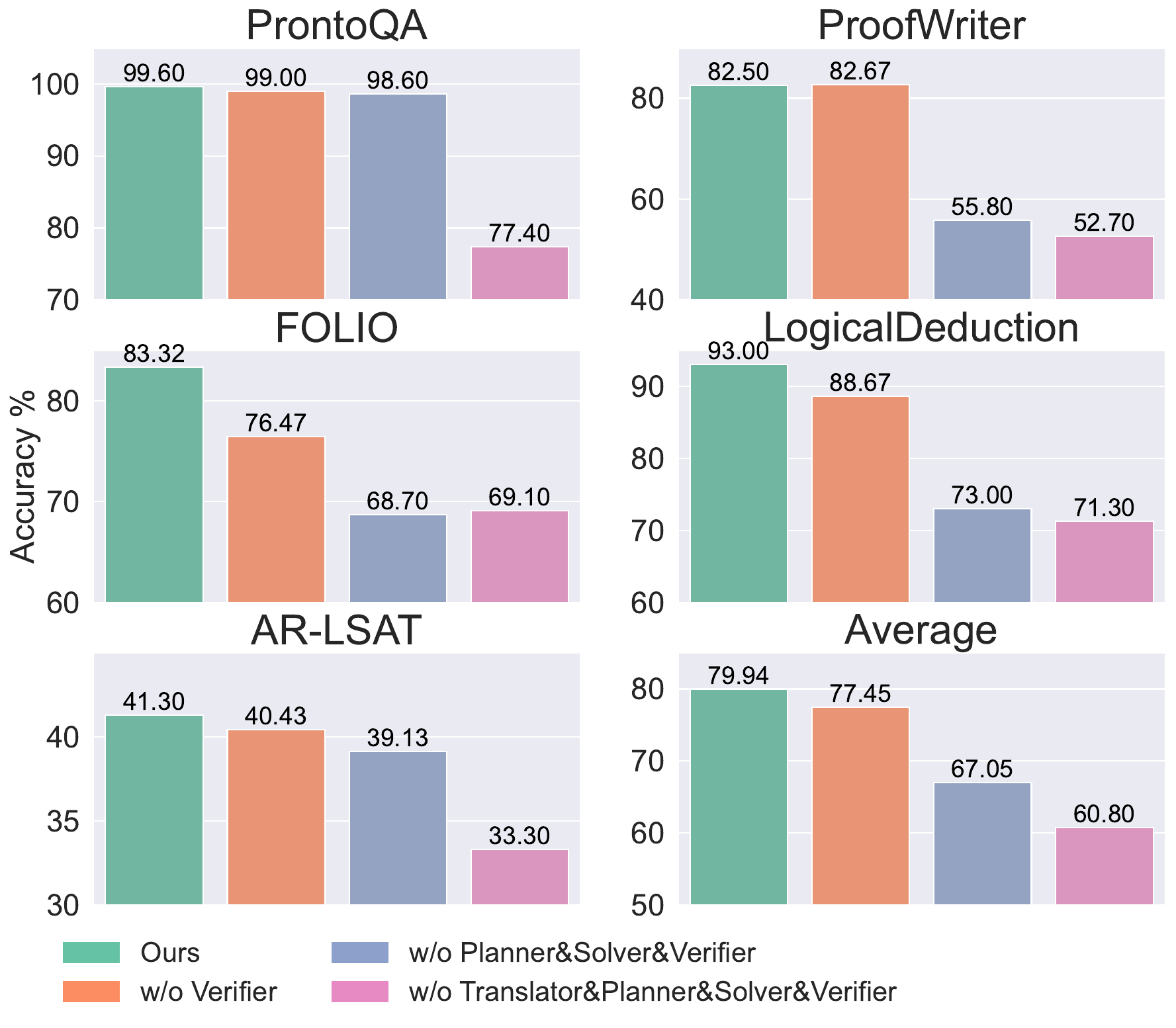}
}
\vspace{-2mm}
\caption{
Ablation study.
Since the Solver is dependent on the Planner, they have to be ablated simultaneously. 
}
\label{Ablation}
\vspace{-4mm}
\end{figure}

\subsection{Model Ablation}
\vspace{-1mm}
To ascertain the individual impact of each module within our framework, we perform an ablation study.
The patterns from Fig. \ref{Ablation} reveal that the contributions to the overall efficacy of our method vary across modules on GPT-4. 
Notably, the Planner and Solver components are identified as the most influential, enhancing performance by an average of 10.4\%, followed by the Translator module, which facilitates a secondary improvement of 6.3\%. 
The finding highlights the efficacy of our proposed \emph{plan-then-solve} design for conquering the raw questions by dividing them into smaller ones. Additionally, the use of symbolic representation and rules shows significant reasoning enhancement.

\begin{figure}[!t]
\centering
\resizebox{0.98\columnwidth}{!}{
  \includegraphics{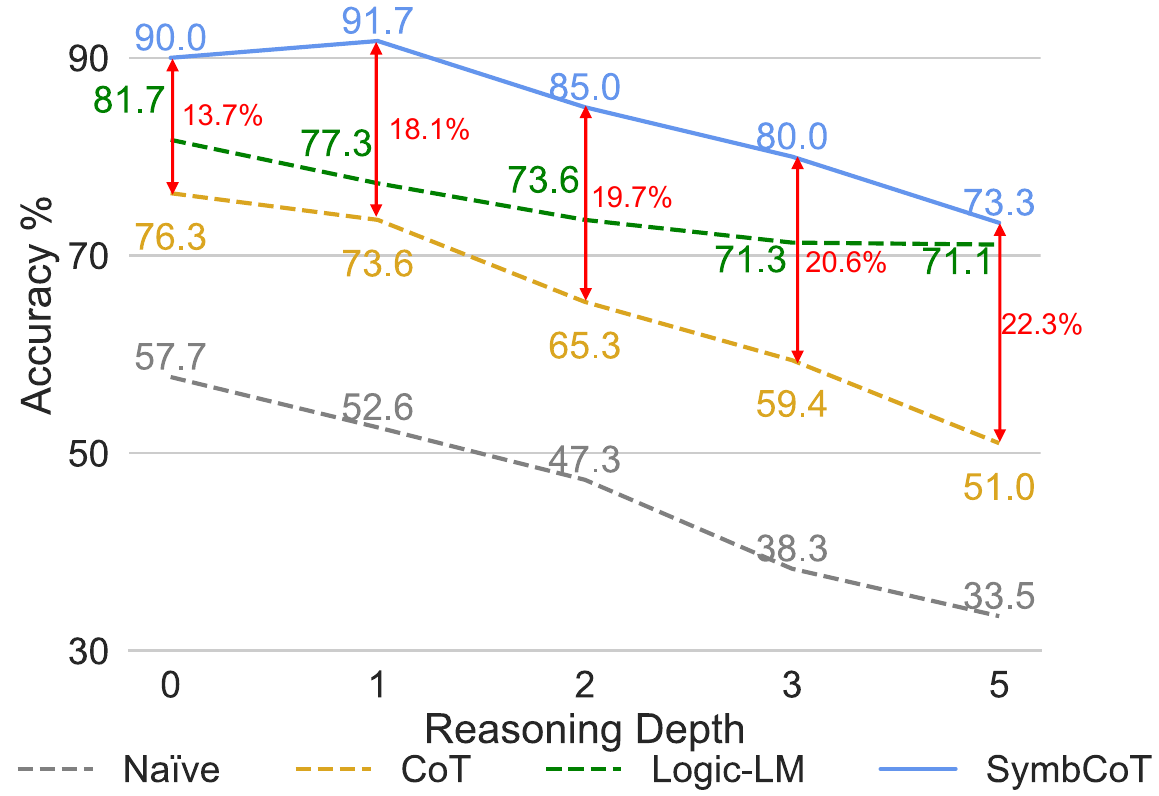}
}
\vspace{-2mm}
\caption{
The effect of reasoning depth with GPT-4 on ProofWriter. 
The red dual-head arrow indicates our improvements over vanilla CoT.
}
\label{depth}
\end{figure}

\begin{figure}[!t]
\centering
\resizebox{0.98\columnwidth}{!}{
  \includegraphics{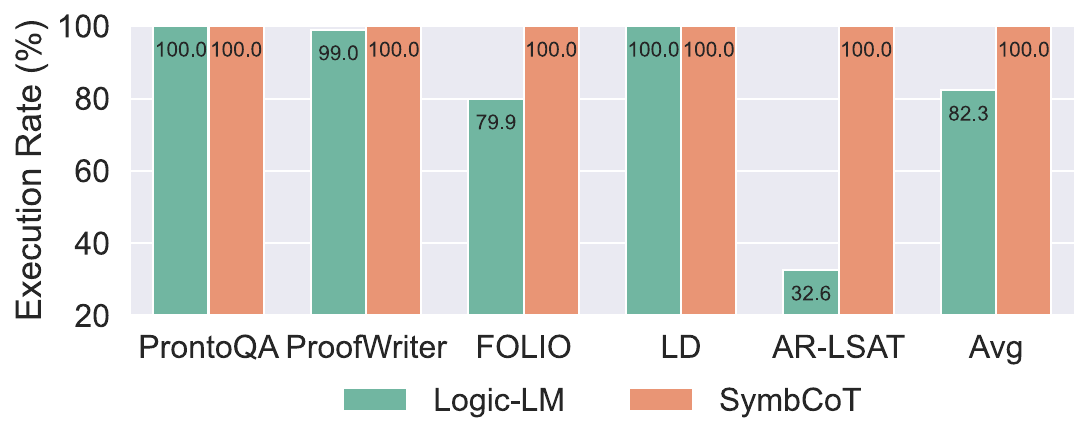}
}
\vspace{-2mm}
\caption{
Execution rate between Logic-LM and Ours.
}
\label{Execution Rate}
\vspace{-4mm}
\end{figure}

\section{Analysis and Discussion}

\vspace{-1mm}
We now delve into our system further and try to explore \emph{why} it advances.

\vspace{-1mm}
\subsection{Performance on Complex Reasoning}

\vspace{-1mm}

In our direct comparison of overall performance, we have demonstrated that our approach surpasses the baseline, particularly noting a significant enhancement in the performance of the CoT. 
Now, we delve deeper into analyzing the performance of different methods across varying levels of reasoning depth. Intuitively, a greater depth indicates more complex problems. 
Observing Fig. \ref{depth}, we notice that as the depth increases, the improvement over CoT becomes more pronounced, suggesting that our advantage lies in tackling more challenging issues. 
Moreover, even at a reasoning depth of 5, our method continues to achieve the best performance.

\subsection{Robustness to Symbolic Syntax Error}

In Fig. \ref{Execution Rate}, we conduct a comparative analysis of our fully LLM-based reasoner against methods that rely on external resolvers, such as Logic-LM, specifically focusing on the success rate of executing symbolic expression syntax. \footnote{For instance, if there are 100 questions and the method successfully executes 80 of them, then the execution rate is 80\%.}
Notably, our method achieves a remarkable execution success rate of up to 100\%. 
This represents a significant improvement over Logic-LM by an average of 17.7\% percentage points. 
Our approach notably enhances the execution rate on the AR-LSAT. It boosts the success rate by 67.4\% from Logic-LM, where LLMs are more prone to translating syntax errors. 
Remarkably, our method consistently executes with 100\% success, showcasing remarkable robustness against syntax errors.

\begin{figure}[!t]
\centering
\resizebox{0.98\columnwidth}{!}{
  \includegraphics{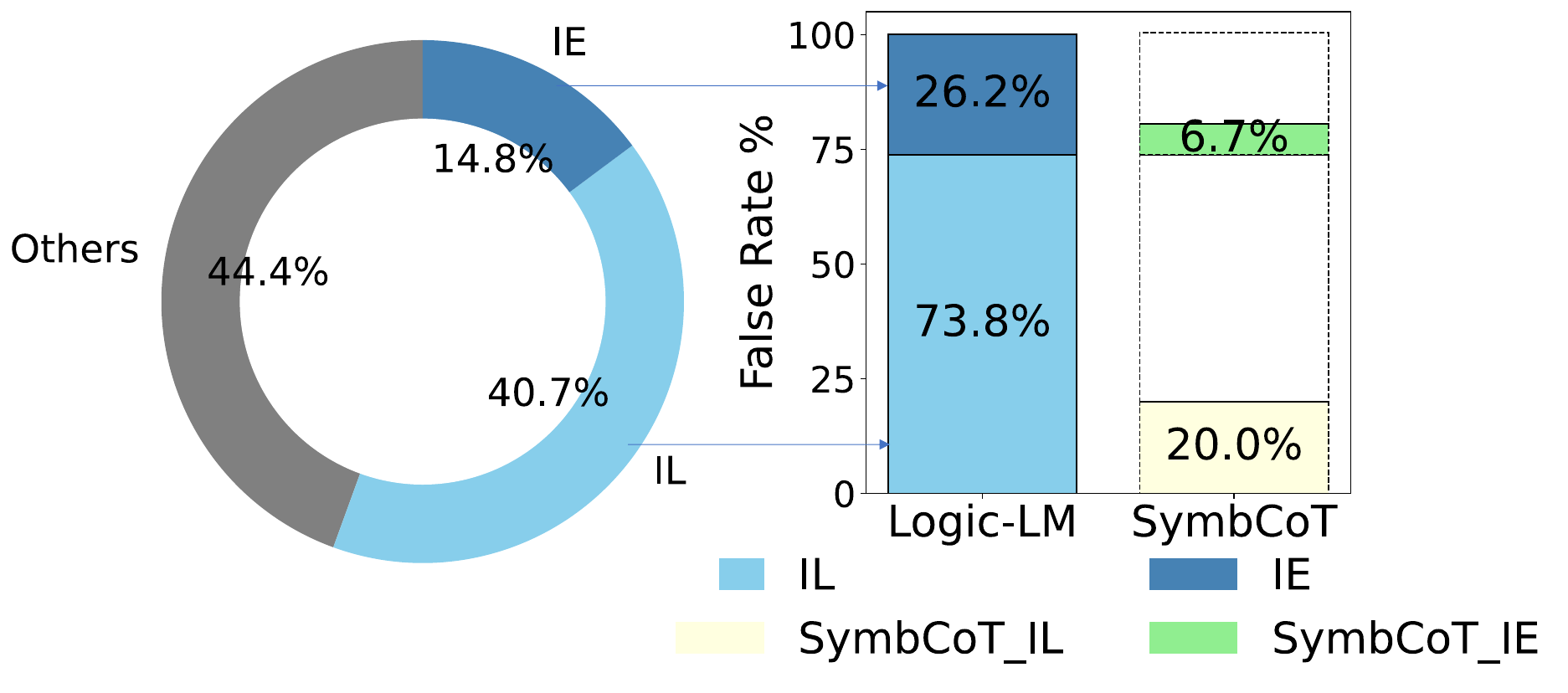}
}
\vspace{-2mm}
\caption{
The left pie shows the error proportion from the external solver due to 1) Information Loss (IL), 2) Information Error (IE), and Others. 
The bar chart consists of two parts. The left bar shows the false rate from the external solver made by IL/IE adding up to 100\%. 
The right bar shows the reduced false rates via our method.
}
\label{Info_loss}
\end{figure}

\begin{figure}[!t]
\centering
\resizebox{0.98\columnwidth}{!}{
  \includegraphics{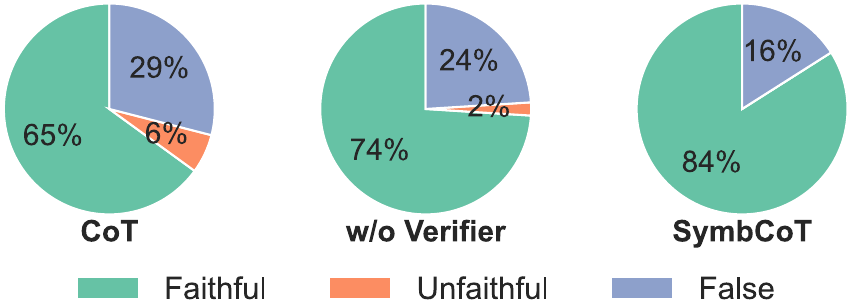}
}
\caption{
The proportion of faithful, unfaithful, and false answers. Faithful/unfaithful denotes whether the predicated correct answer is derived from valid\&reaonsable logical reasoning.
}
\label{Faithfulness}
\vspace{-4mm}
\end{figure}

\vspace{-1mm}

\subsection{Benefit of Hybrid Expression of Symbolic And Natural Language}
\label{section5.3}
\vspace{-1mm}

LLM's translations from natural to symbolic language sometimes omit crucial information or introduce inaccuracies, leading to flawed symbolic representations. 
Our analysis, shown in Fig. \ref{Info_loss}, examines errors in cases wrongfully categorized as `unknown' by external solvers on FOLIO. 
We identify that 55.6\% of these errors were due to information loss (IL, 40.7\%)—where essential details are missed—and information error (IE, 14.8\%)—where translations are incorrect. 
Implementing our methodology reduces these errors by 73.3\%, with significant declines in IL and IE by 53.8\% and 19.5\%, respectively. 
This demonstrates the effectiveness of our LLM-based symbolic reasoning approach, which cross-references both symbolic and natural language data to rectify translation errors and bolster logical reasoning.

\begin{figure}[!t]
\centering
\resizebox{0.98\columnwidth}{!}{
  \includegraphics{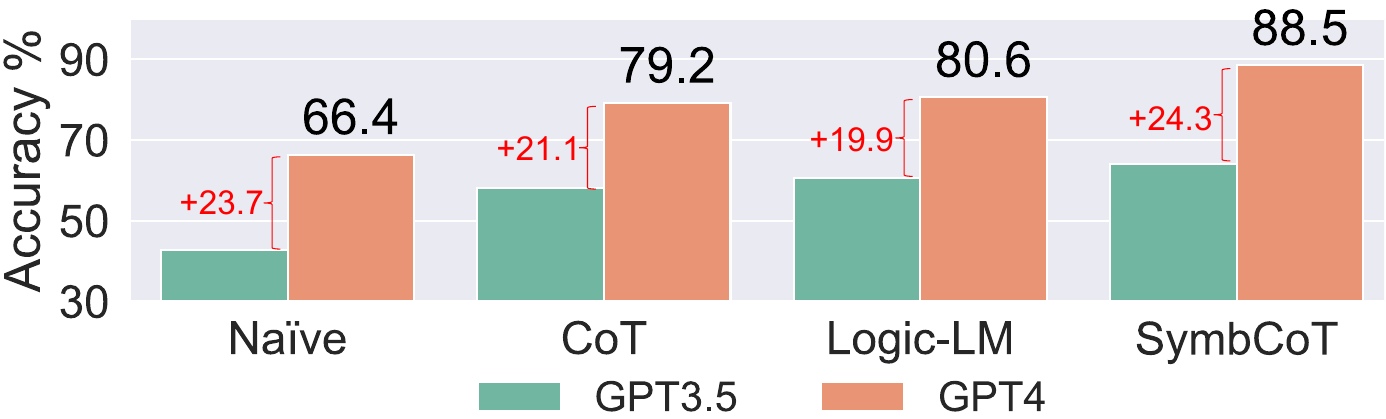}
}
\vspace{-2mm}
\caption{
The Improvement from GPT-3.5 to GPT-4.
}
\label{llm_size}
\vspace{-4mm}
\end{figure}

\subsection{Reasoning Faithfulness}
Often, LLMs may deliver correct answers through flawed reasoning, essentially reaching the right conclusion by luck.
Thus, we further assess the faithfulness of reasoning in the CoT, our SymbCoT, and SymbCoT without a Verifier on the FOLIO dataset in Fig. \ref{Faithfulness}. 
We define an instance as 'faithful' if both the answer and the process are correct and logical; 'unfaithful' if the answer is correct but the process is not; and 'false' if the answer itself is incorrect. 
To verify the logical validity of the reasoning process when the answer is correct, we employed manual evaluation. 
This assessment is carried out by five computer science graduate students with adequate training, and the logical propriety of a process for a given example was determined based on the majority's opinion.
We can see that within the CoT, 6\% of correct answers resulted from flawed reasoning, achieved serendipitously rather than through correct logical steps. 
In SymbCoT without a Verifier, the rate of such unfaithful reasoning dropped to 2\%. 
Integrating a Verifier, we eliminated unfaithful reasoning entirely, showcasing that our approach ensures credible, symbolic-based reasoning and reduces reliance on chance. 
This highlights the effectiveness of our methodology in enhancing the faithfulness of the reasoning process, with the Verifier serving to verify and refine reasoning steps.

\subsection{Impact of Using Different LLMs}
Our comparison of GPT-3.5 and GPT-4 on three FOL datasets (Fig. \ref{llm_size}) shows the most performance boost (24.3\%) with our method upon upgrading models.
This underscores the synergy between our approach and more advanced models. 
In contrast, methods like Logic-LM, which rely on external rule-based solvers, exhibit the least improvements with stronger models due to their dependence on LLMs for translation rather than reasoning. 
Thus, although translation quality may improve, reasoning capabilities remain unchanged as it is capped by external solvers.
Our fully LLM-based method shows significant improvements compared to external solver-based approaches as model capabilities increase, indicating a promising advancement path in line with LLMs' evolution.

\subsection{Error analysis}
The error analysis reveals that for an LLM to accurately complete a reasoning task, it requires two key capabilities:
1) The basic planning ability to identify the correct reasoning path, and
2) Upon identifying the reasoning path, ensure that each step of the reasoning is based on valid logic.
The development of SymbCoT primarily enhances the second capability. However, the first capability, planning, is equally crucial in problem-solving. Consider a scenario where the model fails to identify the correct path; in such instances, there wouldn't even be an opportunity for logical errors to occur, which our method aims to address. Yet, our approach does not enhance the model's planning capabilities, as these are inherently reliant on the LLM itself. Therefore, the LLM needs to meet this reasoning baseline for SymbCoT to be effective.

\subsection{Case Study}
Finally, we showcase two instances to demonstrate in detail the advantages of our method.

\paragraph{CoT makes logical fallacy.}
Our comparison with CoT reveals its susceptibility to logical fallacies, making invalid deductions like affirming the consequent, while our method effectively prevents such errors.

\vspace{-2mm}
{\footnotesize
\begin{enumerate}[label={}, itemsep=0pt, parsep=0pt, leftmargin=*]
    \item \textbf{Premise 1:} There was one six-way tie on the leaderboard and one person in the six-way tie was from Belgium.
    \item \textbf{Premise 2:} Descampe is from Belgium and is on the leaderboard of the 1992 du Maurier Classic.
    \item \textbf{Question Statement:} 
    Descampe is in the six-way tie in the leaderboard of the 1992 du Maurier Classic. (Ground Truth: Unknown)
    \vspace{1.5mm}
    \item \textbf{GPT-4 Incorrect CoT:} The context states that there was one six-way tie on the leaderboard and one person in the six-way tie was from Belgium. Descampe is from Belgium and is on the leaderboard of the 1992 du Maurier Classic. Therefore, it can be inferred that Descampe is in the six-way tie in the leaderboard of the 1992 du Maurier Classic. 
    \vspace{1.5mm}
    \item \textbf{Ours:} From the step from $\exists x (\text{Tie}(x, \text{sixWay}) \land \text{From}(x, \text{Belgium}))$ to Tie(Descampe, sixWay) requires a premise directly stating Descampe's involvement in the six-way tie, which is absent such as Tie(Descampe, sixWay). However, this information is not given. Thus, the truth of the statement "Tie(Descampe, sixWay)" remains unknown.
    \vspace{-1mm}
\end{enumerate}
}

\vspace{-2mm}
\paragraph{External solver fails due to information loss.} 

Using external solvers like Prover9 for FOL entails translation risks, potentially missing key predicates and reducing accuracy. Our LLM-based approach mitigates this by leveraging its natural language comprehension to better capture the context's intricacies.

{\footnotesize 
\begin{enumerate}[label={}, itemsep=0pt, parsep=0pt, leftmargin=*] 
    \item \textbf{Premise 1:} $\forall x (\text{\textit{Book}}(x)\rightarrow\text{\textit{Contains}}(x,\text{knowledge}))$ ::: Books contain tons of knowledge.
    \item \textbf{Premise 2:} $\forall x \forall y (\text{\textit{Person}}(x) \land \text{\textit{Reads}}(x,y)\rightarrow\text{\textit{Gains}}(x,\text{knowledge}))$ ::: When a person reads a book, that person gains knowledge.
    \item \textbf{Premise 3:} $\forall x (\text{\textit{Person}}(x) \land \text{\textit{Gains}}(x,\text{knowledge})\rightarrow\text{\textit{Smarter}}(x))$ ::: If a person gains knowledge, they become smarter.
    \item \textbf{Premise 4:} $\text{\textit{Reads}}(\text{harry},\text{walden})$ ::: Harry read the book “Walden” by Henry Thoreau.
    \item \textbf{Question Statement:}
    Smarter(harry) ::: Harry is smarter than before. (Ground Truth: True) 
    \vspace{1.5mm}
    \item \textbf{Prover9:} Uncertain 
    \vspace{1.5mm}
    \item \textbf{Ours:} Since Walden is a book, and according to the premise $\forall x (\text{Book}(x)\rightarrow\text{Contains}(x, \text{knowledge}))$, it follows that $\text{Book}(\text{walden}) \rightarrow \text{Contains}(\text{walden}, \text{knowledge})$. And Harry reads Walden, which leads to gaining knowledge, as described by the premise $\forall x \forall y (\text{Person}(x) \land \text{Reads}(x, y)\rightarrow\text{Gains}(x, \text{knowledge}))$, thus $\text{Reads}(\text{harry}, \text{walden})\rightarrow\text{Gains}(\text{harry}, \text{knowledge})$. Furthermore, anyone who gains knowledge is considered smarter, as per the premise $\forall x (\text{Person}(x) \land \text{Gains}(x, \text{knowledge})\rightarrow\text{Smarter}(x))$, implying that $\text{Gains}(\text{harry}, \text{knowledge})\rightarrow\text{Smarter}(\text{harry})$.
    \item Thus, Smarter(Harry) is true.
\end{enumerate}
}

\section{Conclusion}
This study presents the Symbolic Chain-of-Thought (SymbCoT) framework, which innovatively integrates symbolic expressions and logical rules with CoT prompting to improve the logical reasoning abilities of LLMs. 
Specifically, based on the backbone LLM, SymbCoT translates the natural language context into the symbolic format and then derives a step-by-step plan to solve the logical reasoning problem with symbolic logical rules, followed by a verifier to check the translation and reasoning chain.
Experimental results on 5 standard benchmarks with both FOL and CO symbolic expressions show that SymbCoT strikingly enhances the vanilla CoT on logical reasoning.

\section*{Acknowledgments}
This work is supported by the Ministry of Education, Singapore, under its MOE AcRF TIER 3 Grant (MOEMOET32022-0001), and also CCF-Baidu Open Fund.

\section*{Limitations}
Our paper has the following potential limitations.
Firstly, we only evaluate two symbolic structures in our framework. 
Despite substantiating the significant potential of First-Order Logic and Constraint Optimization in augmenting the logical reasoning capabilities of LLMs, it is imperative to conduct further assessments on additional symbolic languages to ensure a comprehensive evaluation.
Secondly, the implementation of our framework is associated with considerable expenses. 
This financial implication is attributed to the methodology of incorporating symbolic rules, which inherently involves an extended reasoning chain and, consequently, the generation of an increased number of tokens by the model. 
This escalation in token generation necessitates additional expenditures related to API usage or the allocation of computational resources.

\section*{Ethics Statement}
Our work on the Symbolic Chain-of-Thought (SymbCoT) framework adheres to ethical guidelines emphasizing transparency, fairness, and responsible AI development. We recognize the ethical implications of enhancing LLMs and have conducted our research with a commitment to minimizing biases, ensuring data privacy, and promoting the explainability of AI decisions. 
Our evaluations utilized publicly available or ethically sourced datasets, and we have made efforts to address and mitigate potential biases within these datasets to ensure fairness and objectivity in our findings.

The broader impact of SymbCoT, aimed at improving logical reasoning in LLMs, has the potential to contribute positively to various fields, including natural language processing, knowledge representation, and automated reasoning. 
By introducing a method that combines symbolic expressions with CoT prompting, we foster more faithful, flexible, and explainable logical reasoning capabilities. 
We encourage the responsible use of our findings and technologies, and we commit to ongoing evaluation of our work's societal and ethical implications.

\bibliographystyle{acl_natbib}

\newpage

\appendix

\onecolumn

\section{Future Direction}
Combining SymbCoT with an external solver presents a promising avenue for enhancing our reasoning system. 
SymbCoT's main limitation is sometimes failing to identify the correct reasoning path, while external solvers often struggle with information loss or translation errors. 
As shown in Section \ref{section5.3}, SymbCoT excels at addressing information loss and correcting errors, while the exhaustive, broad-spectrum search nature of the external solver is more effective in identifying reasoning paths.
Therefore, these methods possess the potential for synergy.

Future work will focus on developing a framework that integrates SymbCoT with an external solver, leveraging their complementary strengths. 
We aim to optimize their interaction to improve overall performance, conducting experiments to validate and refine this hybrid approach.

\section{Symbolic Representation}
\label{sec:symbolic_specific}
\subsection{First-Order Logic}
First-Order Logic, also known as predicate logic or first-order predicate calculus, is a formal system used in mathematics, philosophy, linguistics, and computer science to express statements about objects and their relationships. 
It extends propositional logic by introducing quantified variables that can represent objects in a domain of discourse and predicates that can express properties of objects and relations between them. 

The syntax of First-Order Logic (FOL) provides a formal structure for constructing expressions that can represent statements about objects, their properties, and the relationships between them. It defines the rules for forming well-formed formulas (WFFs), which are expressions in FOL that can be evaluated as true or false. Below, we introduce the key components of FOL syntax:

\paragraph{Syntax.}

The syntax of FOL consists of a set of symbols divided into several categories:
\begin{itemize}
    \item \textbf{Variables:} Symbols used to stand for objects in the domain of discourse (e.g., \(x, y, z\)).
    \item \textbf{Constant symbols:} Names that refer to specific objects in the domain (e.g., \(a, b, c\)).
    \item \textbf{Predicate symbols:} Symbols that represent properties of objects or relations between objects, with an associated arity (e.g., \(P, Q, R\)).
    \item \textbf{Function symbols:} Symbols representing functions from objects to objects, also with arity (e.g., \(f, g, h\)).
    \item \textbf{Logical connectives:} Symbols used to combine or modify statements, including \(\land, \lor, \neg, \rightarrow, \leftrightarrow\).
    \item \textbf{Quantifiers:} Symbols expressing quantification over objects, including the universal quantifier (\(\forall\)) and the existential quantifier (\(\exists\)).
    \item \textbf{Parentheses and commas:} Used for grouping and separating arguments.
\end{itemize}

\paragraph{Terms}

Terms are the simplest elements in FOL that refer to objects in the domain. Examples include:
\begin{itemize}
    \item A variable: \(x\)
    \item A constant: \(a\)
    \item A function applied to terms: \(f(a, x)\)
\end{itemize}

\paragraph{Atomic Formulas (or Atoms)}

Atomic formulas are formed by applying a predicate symbol to the appropriate number of terms. An example of an atomic formula is:
\[P(x, a)\]
where \(P\) is a predicate symbol, \(x\) is a variable, and \(a\) is a constant.

\paragraph{Complex Formulas}

Complex formulas are constructed from atomic formulas using logical connectives and quantifiers. For example:
\[\forall x (P(x) \rightarrow Q(f(x)))\]
This complex formula states that for all \(x\), if \(P(x)\) is true, then \(Q(f(x))\) is also true, where \(Q\) is another predicate symbol and \(f\) is a function symbol.

\paragraph{Quantification}

Quantifiers are used to express statements about all objects or the existence of objects in the domain. Examples include:
\begin{itemize}
    \item Universal quantification: \(\forall x P(x)\), stating that \(P\) holds for all \(x\).
    \item Existential quantification: \(\exists x P(x)\), stating that there exists an \(x\) for which \(P\) holds.
\end{itemize}

\paragraph{Well-formed Formulas (WFFs)}

Expressions that adhere to the syntactic rules of FOL are considered well-formed. These expressions include terms, atomic formulas, and complex formulas constructed according to the rules of formation. The syntax ensures that the expressions are meaningful within the logic.

\subsection{Constraint Optimization}

Constraint optimization is a branch of optimization theory that focuses on finding the most favorable solution from a set of possible options, subject to a series of constraints. It is widely used in various fields such as operations research, artificial intelligence, engineering, and economics to solve problems that involve limitations or restrictions on the possible solutions.

\paragraph{Definition.}

Constraint optimization involves determining the optimal solution(s) to a problem defined by:
\begin{itemize}
    \item A set of variables, each of which can take on values from a given domain.
    \item A set of constraints that specify allowable combinations of values for subsets of variables.
    \item An objective function that needs to be maximized or minimized.
\end{itemize}

\paragraph{Components}

The key components of a constraint optimization problem include:
\begin{itemize}
    \item \textbf{Variables:} The elements that represent the aspects of the problem being optimized (e.g., \(x, y, z\)).
    \item \textbf{Domains:} The possible values that each variable can take (e.g., \(D(x), D(y), D(z)\)).
    \item \textbf{Constraints:} Conditions or relations that must be satisfied by the variables (e.g., \(x + y \leq z\)).
    \item \textbf{Objective Function:} A mathematical expression that defines the goal of the optimization, which can be either maximization or minimization (e.g., \(f(x, y, z) = x^2 + y^2 + z^2\)).
\end{itemize}

\section{Dataset Statistics}
\label{dataset_specific}
In alignment with the established norms of prior research, notably, the practices adopted by Logic-LM \cite{pan-etal-2023-logic}, our study employs identical records for the sake of ensuring consistency and facilitating comparative analysis. Table \ref{Dataset_stats} shows the statistics of each dataset.

\begin{table}[h]
\centering
\captionsetup{justification=centering, margin=2cm}
\label{tab:dataset_stats}
\begin{tabular}{@{}lccc@{}}
\toprule
\begin{tabular}[c]{@{}l@{}}Dataset\end{tabular} & \begin{tabular}[c]{@{}c@{}}Symbolic\\ Structure\end{tabular} & \begin{tabular}[c]{@{}c@{}}Test\\ Size\end{tabular} & \begin{tabular}[c]{@{}c@{}}Opts\\ Num\end{tabular} \\ \midrule
PrOntoQA           & FOL      & 500       & 2                 \\
ProofWriter        & FOL      & 600       & 3                 \\
FOLIO              & FOL            & 204       & 3                 \\
LogicalDeduction   & CSP            & 300       & 3, 5, 7           \\
AR-LSAT            & CSP             & 230       & 5                 \\ \bottomrule
\end{tabular}
\caption{Dataset Statistics}
\label{Dataset_stats}
\end{table}

\section{Baseline Specification}
\label{baseline_specific}

Here are the details of each baseline:
\paragraph{Naive Prompting.} This approach involves instructing the LLMs to directly answer questions with the aid of a few in-context demonstrations, prioritizing straightforward response generation without explicit reasoning processes; 

\paragraph{Chain-of-Thought (CoT).} CoT encourages the LLM to articulate a step-by-step reasoning process leading to the final answer, supported by in-context demonstrations \cite{cot};

\paragraph{Logic-LM} Distinguished by its utilization of the LLM to translate natural language contexts and questions into a symbolic representation, this baseline leverages external solvers to deduce answers, offering a hybrid approach to problem-solving \cite{pan-etal-2023-logic}.

\paragraph{CoT-SC} An advanced version of CoT, which employs self-consistency by an ensemble strategy through majority-voting among multiple CoTs \cite{CoT-SC};

\paragraph{Tree-of-thought (ToT)} This approach enables LLMs to explore multiple reasoning pathways through a decision-making tree structure, incorporating backtracking via a search algorithm to identify the globally optimal reasoning path \cite{ToT};

\paragraph{Cummulative Reasoning (CR)} CR utilizes the LLMs in a cumulative and iterative manner, mimicking the human thought process by building upon previous reasoning steps to gradually approach the solution \cite{CR};

\paragraph{DetermLR} This method constructs a logical reasoning process, transitioning from indeterminate premises to a determinate conclusion iteratively, and integrates external modules such as memory storage to support this process \cite{DetermLR}.

\noindent To ensure efficient resource management, we adopt the experimental data from Logic-LM \cite{pan-etal-2023-logic} and DetermLR \cite{DetermLR} for comparative analysis as the datasets and records we used on our experiments are exactly the same. The primary metric for evaluation is accuracy, reflecting the multiple-choice format of the questions across the datasets.

\section{Prediction Behavior}

\begin{figure}[!t]
\centering
\includegraphics[width=0.7\columnwidth]{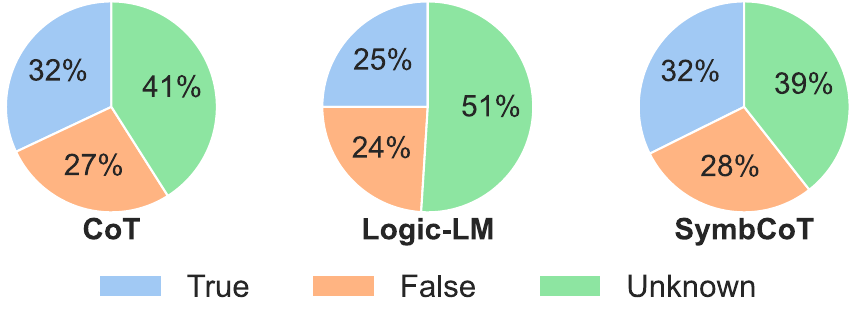}
\vspace{-2mm}
\caption{
The proportion of predicting True, False, and Unknown on FOLIO.
}
\label{prediction_behaviour}
\vspace{-4mm}
\end{figure}

We present the predicting behavior of Chain-of-Thought (CoT), Logic-LM, and our method by showing the percentage of predicting each option in Fig. \ref{prediction_behaviour}. 
Please note that for Logic-LM, to see the behavior of the external solver (Prover9), we turn off their backup strategy which uses the result of CoT when the records containing syntax error cannot be execute by the external solver.
We found that Logic-LM, using an external solver, tends to predict more "Unknown". This could be attributed to the loss of information during the translation process, which results in insufficient data for accurately deducing the correct answers. In such instances, these tools opt to predict an outcome as "unknown", where answers that are actually "True" or "False" as a consequence of this information loss.

\section{Evaluation Merics}

\begin{figure}[!t]
\centering
\includegraphics[width=0.7\columnwidth]{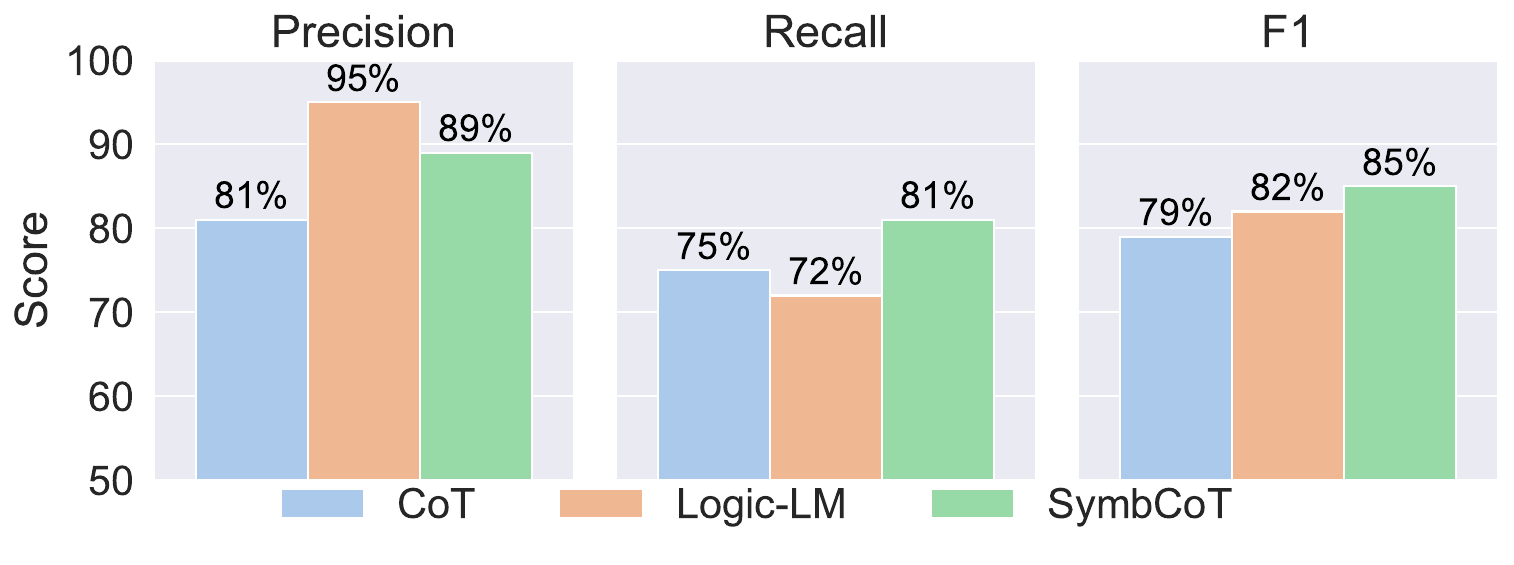}
\vspace{-2mm}
\caption{
The Precision, Recall, and F1 score for CoT, Logic-LM and Ours on FOLIO.
}
\label{evaluation matrix}
\vspace{-4mm}
\end{figure}

In Fig. \ref{evaluation matrix}, the data encapsulates a comparative analysis of performance metrics—precision, recall, and F1 score—across three models: CoT (Chain of Thought), Logic-LM, and our proposed method.
Ours method surpass others on the F1 score, by 6\% on CoT and 3\% on Logic-LM. 

\subsection{F1 Score Analysis}
Starting with the F1 scores, which balance precision and recall, providing a measure of a model's accuracy while considering both the false positives and false negatives. 
Our model achieves the highest F1 score of 85\%, indicating a strong balance between precision and recall. 
This suggests that our approach effectively mitigates the trade-offs between these two metrics, potentially offering a more balanced and reliable model for various applications.

In comparison, Logic-LM posts an F1 score of 82. While slightly lower than ours, this score still represents a commendable balance between precision and recall, hinting at the effectiveness of incorporating logical reasoning within the model's framework. However, it slightly lags behind our model, possibly due to differences in handling information or reasoning processes.

CoT model scores the lowest in terms of F1, at 79. This indicates a more considerable discrepancy between precision and recall compared to the other models. The CoT approach, while innovative, may face challenges in optimally balancing false positives and false negatives, which is reflected in its overall performance metric.

\subsection{Recall Analysis}
When looking at recall, our model scores the highest of 81, highlighting its capability to identify a high percentage of all relevant instances. 
This performance is crucial in scenarios where missing a positive case can have significant implications, such as medical diagnoses or fraud detection.

Logic-LM has a recall rate of 72, which is the lowest among the three models. 
This suggests that while Logic-LM might be precise, it potentially misses a larger portion of relevant instances, a limitation that could impact its utility in applications where high recall is critical.

CoT exhibits a recall of 75, positioning it between our methodology and Logic-LM. This indicates a moderate ability to capture relevant instances but also room for improvement, especially when compared to our method.

\subsection{Precision Analysis}
Considering precision, our model demonstrates a score of 89. 
This high precision indicates that the majority of the instances it identifies as positive are indeed positive, which is essential for reducing the cost of false positives in many applications.

Logic-LM achieves the highest precision at 95, underscoring its strength in accurately predicting positive instances when it does make a prediction. 
This high precision is particularly valuable in contexts where the cost of false positives is high, even though it may sacrifice some recall to achieve this.

CoT has a precision of 81, the lowest among the three. 
While still respectable, this suggests that CoT may generate more false positives than Logic-LM and ours, affecting its efficiency and reliability in certain contexts.

\subsection{Conclusion on Evaluation Metrics}
In summary, the analysis reveals that our methodology offers a balanced performance with the highest F1 score, suggesting it effectively balances precision and recall. Logic-LM stands out for its precision, making it suitable for applications where accuracy is paramount, at the expense of recall. 
CoT, while showing promising results, trails in both precision and recall, indicating potential areas for improvement. 
Each model has its strengths and potential applications, with our approach showing a well-rounded performance that could be particularly beneficial in scenarios requiring a balance between identifying relevant instances and maintaining high accuracy.

\section{Output Average Length}
Table \ref{tab:output_length} is the average number of output tokens for each module and dataset.

\begin{table}[h]
\centering
\begin{tabular}{lccccc}
\toprule
\textbf{Model} & \textbf{ProntoQA} & \textbf{ProofWriter} & \textbf{FOLIO} & \textbf{LogicalDeduction} & \textbf{AR-LSAT} \\
\midrule
\textbf{Translator} & 324 & 290 & 145 & 176 & 323 \\
\textbf{Planner}    & 198 & 202 & 313 & 251 & 627 \\
\textbf{Solver}     & 214 & 216 & 330 & 282 & 654 \\
\textbf{Verifier}   & 384 & 353 & 334 & 472 & 726 \\
\bottomrule
\end{tabular}
\caption{Output length across different datasets and modules}
\label{tab:output_length}
\end{table}

\section{Output examples}

To have a better understanding of how SymnbCoT solves logical questions, we select a typical solving example below that showcases the step-by-step logical deduction.

\begin{tcolorbox}[breakable,colback=gray!5,colframe=nmgray!75!black,before skip=1pt, after skip=1pt,fontupper=\linespread{0.8}\selectfont]
{\footnotesize

\textbf{Input:}

Predicates:
\begin{itemize}
    \item Rabbit(x) ::: x is a rabbit.
    \item Cute(x) ::: x is cute.
    \item Turtle(x) ::: x is a turtle.
    \item Animal(x) ::: x is an animal.
    \item Squirrel(x) ::: x is a squirrel.
    \item Skittish(x) ::: x is skittish.
    \item Still(x) ::: x is still.
\end{itemize}

Premises:
\begin{itemize}
    \item $\forall x (\text{Rabbit}(x) \rightarrow \text{Cute}(x))$ ::: All rabbits are cute.
    \item $\exists x \, \text{Turtle}(x)$ ::: Some turtles exist.
    \item $\forall x (\text{Animal}(x) \rightarrow (\text{Rabbit}(x) \oplus \text{Squirrel}(x)))$ ::: An animal is either a rabbit or a squirrel.
    \item $\forall x (\text{Skittish}(x) \rightarrow \neg \text{Still}(x))$ ::: If something is skittish, then it is not still.
    \item $\forall x (\text{Squirrel}(x) \rightarrow \text{Skittish}(x))$ ::: All squirrels are skittish.
    \item $\text{Still}(\text{rock})$ ::: Rock is still.
\end{itemize}

Conclusion:
\begin{itemize}
    \item $(\text{Cute}(\text{rock}) \land \text{Still}(\text{rock})) \rightarrow (\text{Turtle}(\text{rock}) \land \text{Skittish}(\text{rock}))$ ::: If Rock is cute and still, then Rock is a turtle and skittish.
\end{itemize}

\textbf{Output:}
\begin{itemize}
    \item \textbf{Step 1: Identify the Goal}\\
    Our goal is to determine the truth of the statement \texttt{"(Cute(rock) $\land$ Still(rock)) $\to$ (Turtle(rock) $\land$ Skittish(rock))"}. This requires verifying if the conditions of Rock being cute and still necessarily lead to Rock being a turtle and skittish.

    \item \textbf{Step 2: Utilize Premise 6 (Still(rock))}\\
    Premise: \texttt{Still(rock)}\\
    Inference Rule: Existential Instantiation\\
    Logical Process: We instantiate Rock and confirm it is still.

    \item \textbf{Step 3: Apply Premise 4 (Skittish(x) $\to$ ¬Still(x))}\\
    Premise: \(\forall x (\text{Skittish}(x) \rightarrow \neg \text{Still}(x))\)\\
    Inference Rule: Universal Instantiation and Modus Tollens\\
    Logical Process: If an entity \(x\) is skittish, then it is not still. Since Rock is still, it cannot be skittish.

    \item \textbf{Step 4: Utilize Premise 5 (Squirrel(x) $\to$ Skittish(x))}\\
    Premise: \(\forall x (\text{Squirrel}(x) \rightarrow \text{Skittish}(x))\)\\
    Inference Rule: Universal Instantiation and Modus Tollens\\
    Logical Process: If an entity \(x\) is a squirrel, then it is skittish. Since Rock is not skittish, it cannot be a squirrel.

    \item \textbf{Step 5: Apply Premise 3 (Animal(x) $\to$ (Rabbit(x) $\oplus$ Squirrel(x)))}\\
    Premise: \(\forall x (\text{Animal}(x) \rightarrow (\text{Rabbit}(x) \oplus \text{Squirrel}(x)))\)\\
    Inference Rule: Universal Instantiation and Modus Tollens\\
    Logical Process: If an entity \(x\) is an animal, then it is either a rabbit or a squirrel. Since Rock is not a squirrel, if it is an animal, it must be a rabbit.

    \item \textbf{Step 6: Combine the Steps}\\
    Logical Process: Combining the above steps, we have established that Rock is still, not skittish, not a squirrel, and if it is an animal, it must be a rabbit. Therefore, Rock cannot be a turtle and skittish at the same time.

    \item \textbf{Step 7: Conclude}\\
    Conclusion: Since we have found that Rock cannot be a turtle and skittish at the same time, the statement \texttt{"(Cute(rock) $\land$ Still(rock)) $\to$ (Turtle(rock) $\land$ Skittish(rock))"} is False. Throughout the process, we used logical operators such as \(\land\) (conjunction), \(\rightarrow\) (implication), and \(\exists\) (existential), and inference rules like Modus Ponens, Universal Instantiation, and Modus Tollens. This logical deduction confirms the falsity of the question statement.\\
    \textbf{Final answer:} \{false\}
\end{itemize}

}
\end{tcolorbox}


\section{Full Prompting}
\label{prompt_specific}

Below is the detailed prompt of our method across all 5 datasets:
\subsection{Ours}
\subsubsection{ProntoQA}

\begin{tcolorbox}[breakable, title=Step-1: Translating natural language context into symbolic,colback=gray!5,colframe=nmgray!75!black,before skip=1pt, after skip=1pt,fontupper=\linespread{0.8}\selectfont]
{\footnotesize

\textbf{Task Description:}

You are given a problem description and a question. The task is to:
\begin{enumerate}
    \item Define all the predicates in the problem.
    \item Parse the problem into logic rules based on the defined predicates.
    \item Write all the facts mentioned in the problem.
    \item Parse the question into the logic form.
\end{enumerate}

\textbf{Problem:}
\begin{itemize}[leftmargin=*]
    \item Each jompus is fruity. Every jompus is a wumpus. Every wumpus is not transparent. Wumpuses are tumpuses. Tumpuses are mean. Tumpuses are vumpuses. Every vumpus is cold. Each vumpus is a yumpus. Yumpuses are orange. Yumpuses are numpuses. Numpuses are dull. Each numpus is a dumpus. Every dumpus is not shy. Impuses are shy. Dumpuses are rompuses. Each rompus is liquid. Rompuses are zumpuses. Alex is a tumpus.
\end{itemize}

\textbf{Question:}
\begin{itemize}[leftmargin=*]
    \item True or false: Alex is not shy.
\end{itemize}

\textbf{Predicates:}
\begin{itemize}[leftmargin=*]
    \item \texttt{Jompus(x)}: Is \textit{x} a jompus?
    \item \texttt{Fruity(x)}: Is \textit{x} fruity?
    \item (... More predicates ...)
\end{itemize}

\textbf{Facts:}
\begin{itemize}[leftmargin=*]
    \item \texttt{Tumpuses(Alex)}: Alex is a tumpus. 
    \item (... More facts ...)
\end{itemize}

\textbf{Rules:}
\begin{itemize}[leftmargin=*]
    \item \texttt{Jompus(x) $\Rightarrow$ Fruity(x)}: Each jompus is fruity.
    \item (... More rules ...)
\end{itemize}

\textbf{Query:}
\begin{itemize}[leftmargin=*]
    \item Shy(Alex, False) ::: Alex is not shy
\end{itemize}

}
\end{tcolorbox}

\begin{tcolorbox}[breakable, title=Step-2: Deriving plan based on the natural and symbolic context,colback=gray!5,colframe=nmgray!75!black,before skip=1pt, after skip=1pt,fontupper=\linespread{0.8}\selectfont]
{\footnotesize

\textbf{Task:}

Can you derive a step-by-step plan that can use the premises and First-Order Logic inference rules to infer new knowledge if needed, in order to prove the statement? Start with identifying the goal and break down the necessary logical inference step by step.

\textbf{First-Order Logic Inference Rules:}

The First-Order Logic inference rules include but are not limited to: Modus Ponens, Modus Tollens, (... More FOL rules ...).

\textbf{Predicates:}
\begin{itemize}[leftmargin=*]
    \item \texttt{Jompus(x)}: Is \textit{x} a jompus?
    \item \texttt{Fruity(x)}: Is \textit{x} fruity?
    \item (... More predicates ...)
\end{itemize}

\textbf{Facts:}
\begin{itemize}[leftmargin=*]
    \item \texttt{Tumpuses(Alex)}: Alex is a tumpus. 
    \item (... More facts ...)
\end{itemize}

\textbf{Rules:}
\begin{itemize}[leftmargin=*]
    \item \texttt{Jompus(x) $\Rightarrow$ Fruity(x)}: Each jompus is fruity.
    \item (... More rules ...)
\end{itemize}

\textbf{Query:}
\begin{itemize}[leftmargin=*]
    \item Shy(Alex, False) ::: Alex is not shy
\end{itemize}

\textbf{Plan:}

To determine whether the statement "Shy(Alex, False)" is true, false, or unknown based on the given premises, we will use a systematic approach applying First-Order Logic inference rules. Here's the step-by-step plan:
\begin{itemize}[leftmargin=*]
    \item 1. Identify the Goal: Our goal is to establish whether Alex is not shy (Shy(Alex, False))

    \item 2. Utilize Facts about Alex: We have Tumpuses(Alex, True) which states Alex belongs to Tumpuses.

    \item (... More plan ...)

    \item By following the plan, you can determine whether "Shy(Alex, False)" is true or false.
\end{itemize}

}

\end{tcolorbox}

\begin{tcolorbox}[breakable, title=Step-3: Solving the Problem given the context and plan,colback=gray!5,colframe=nmgray!75!black,before skip=1pt, after skip=1pt,fontupper=\linespread{0.8}\selectfont]
{\footnotesize

\textbf{Task:}

Based on the premises, question and plan, execute each step by selecting relevant premises and making inference based on First-order Logic inference rules to solve the question. Ensure to follow the plan and indicate what First-Order Logic inference rules you used.

\textbf{First-Order Logic Inference Rules:}

The First-Order Logic inference rules include but are not limited to: Modus Ponens, Modus Tollens, (... More FOL rules ...)

\textbf{Predicates:}
\begin{itemize}[leftmargin=*]
    \item \texttt{Jompus(x)}: Is \textit{x} a jompus?
    \item \texttt{Fruity(x)}: Is \textit{x} fruity?
    \item (... More predicates ...)
\end{itemize}

\textbf{Facts:}
\begin{itemize}[leftmargin=*]
    \item \texttt{Tumpuses(Alex)}: Alex is a tumpus. 
    \item (... More facts ...)
\end{itemize}

\textbf{Rules:}
\begin{itemize}[leftmargin=*]
    \item \texttt{Jompus(x) $\Rightarrow$ Fruity(x)}: Each jompus is fruity.
    \item (... More rules ...)
\end{itemize}

\textbf{Query:}
\begin{itemize}[leftmargin=*]
    \item Shy(Alex, False) ::: Alex is not shy
\end{itemize}

\textbf{Plan:}

To determine whether the statement "Shy(Alex, False)" is true, false, or unknown based on the given premises, we will use a systematic approach applying First-Order Logic inference rules. Here's the step-by-step plan:
\begin{itemize}[leftmargin=*]
    \item 1. Identify the Goal: Our goal is to establish whether Alex is not shy (Shy(Alex, False))

    \item 2. Utilize Facts about Alex: We have Tumpuses(Alex, True) which states Alex belongs to Tumpuses.

    \item (... More plan ...)

    \item By following the plan, you can determine whether "Shy(Alex, False)" is true or false.
\end{itemize}

\textbf{Execution:}

Let's execute the plan step by step, applying First-Order Logic inference rules.

\begin{itemize}
    \item Identify the Goal: Goal: Determine if \text{Shy(Alex, False)} \text{ is true.}

    \item Utilize Facts about Alex: 
        - \text{Fact: } \text{Tumpuses(Alex, True)}.  
       - \text{Inference Rule: Given Fact.}  
       - \text{Logical Process: } \text{Tumpuses(Alex, True)} \text{ is given.}

    \item (... More solving steps ...)

    \item Thus, "Shy(Alex, False)" is true based on the logical deductions.
\end{itemize}
}
\end{tcolorbox}

\begin{tcolorbox}[breakable, title=Step-4: Verifying the translation and solving process ,colback=gray!5,colframe=nmgray!75!black,before skip=1pt, after skip=1pt,fontupper=\linespread{0.8}\selectfont]
{\footnotesize

\textbf{Task:}

The task is to verify whether the original execution correctly determines the value of the given conclusion based on the context.
You should abide by the First-Order Logic rules when checking the original execution and the conclusion. \\
When verifying the logical process and the final answer, make sure you fully consider the following aspects before verification: \\
1. Whether the logical process abides by the First-Order Logic inference rules. \\
2. Assumption and Consistency Verification. Please check whether the relevant facts or rules used in a logical step are indeed from the context or inferred from the previous steps. If you think it contains assumption in the logical steps, you need to clearly what assumption it is and verify whether the assumptions are from the context and whether it is already inferred from the previous steps. Please note that an assumption that does not exist from the context or was not inferred from the previous step will be considered an invalid assumption.
Please note that if an assumption is semantically equivalent to context or information inferred from the context, it should be seen as a valid assumption. For example, "X sees X" can be interpreted as "X sees itself" since they are semantically equivalent. \\
3. Whether the translation of the First-Order Logic for context and question is semantically consistent with the original natural language. \\
4. Refinement of Logical Process: If the original execution violates any of the above steps, refine the logical process. Use the relevant premises and information correctly derived from previous steps to create a more accurate logical deduction. \\
5. If applicable, please provide a detailed analysis of each step in the logical process and a refined answer at the end, using the format {true/false}. The answer should only be true or false but NOT unknown.
6. Make sure the conclusion aligns with the information inferred from the logical steps. \\

\textbf{Predicates:}
\begin{itemize}[leftmargin=*]
    \item \texttt{Jompus(x)}: Is \textit{x} a jompus?
    \item \texttt{Fruity(x)}: Is \textit{x} fruity?
    \item (... More predicates ...)
\end{itemize}

\textbf{Facts:}
\begin{itemize}[leftmargin=*]
    \item \texttt{Tumpuses(Alex)}: Alex is a tumpus. 
    \item (... More facts ...)
\end{itemize}

\textbf{Rules:}
\begin{itemize}[leftmargin=*]
    \item \texttt{Jompus(x) $\Rightarrow$ Fruity(x)}: Each jompus is fruity.
    \item (... More rules ...)
\end{itemize}

\textbf{Query:}
\begin{itemize}[leftmargin=*]
    \item Shy(Alex, False) ::: Alex is not shy
\end{itemize}

\textbf{Original Execution:}

Let's execute the plan step by step, applying First-Order Logic inference rules.

\begin{itemize}
    \item Identify the Goal: Goal: Determine if \text{Shy(Alex, False)} \text{ is true.}

    \item Utilize Facts about Alex: 
        - \text{Fact: } \text{Tumpuses(Alex, True)}.  
       - \text{Inference Rule: Given Fact.}  
       - \text{Logical Process: } \text{Tumpuses(Alex, True)} \text{ is given.}

    \item (... More solving steps ...)

    \item Thus, "Shy(Alex, False)" is true based on the logical deductions.
\end{itemize}

\textbf{Verification:}

To verify the logical validity of the original execution concerning the statement "Shy(Alex, False)" based on the given premises, let's assess each step against the requirements of First-Order Logic, assumptions, consistency, and semantic equivalence.\\
\textbf{Context verification:}
\begin{itemize}
    
    \item  The context and query correctly correspond to the information in the original context and are semantically consistent.

\end{itemize}

\textbf{Logical Process Verification:}
\begin{itemize}
    \item 1. Fact: Tumpuses(Alex, True) is directly from the context with no invalid assumption.
    \item (... More logical verification ...)
    \item Therefore, after verifying the translation between the original context and symbolic format, and the logical process, the original conclusion "Shy(Alex, Flase) is true" is valid and remains unchanged. 
\end{itemize}
}
\end{tcolorbox}

\subsubsection{ProofWriter}

\begin{tcolorbox}[breakable, title=Step-1: Translating natural language context into symbolic,colback=gray!5,colframe=nmgray!75!black,before skip=1pt, after skip=1pt,fontupper=\linespread{0.8}\selectfont]
{\footnotesize

\textbf{Task Description:}

You are given a problem description and a question. The task is to:
\begin{enumerate}
    \item Define all the predicates in the problem.
    \item Parse the problem into logic rules based on the defined predicates.
    \item Write all the facts mentioned in the problem.
    \item Parse the question into the logic form.
\end{enumerate}

\textbf{Problem:}
\begin{itemize}[leftmargin=*]
    \item Anne is quiet. Erin is furry. Erin is green. Fiona is furry. Fiona is quiet. Fiona is red. Fiona is rough. Fiona is white. Harry is furry. Harry is quiet. Harry is white. Young people are furry. If Anne is quiet then Anne is red. Young, green people are rough. If someone is green then they are white. If someone is furry and quiet then they are white. If someone is young and white then they are rough. All red people are young.
\end{itemize}

\textbf{Question:}
\begin{itemize}[leftmargin=*]
    \item Is the following statement true, false, or unknown? Anne is white.
\end{itemize}

\textbf{Translation: }\\
\textbf{Predicates:}
\begin{itemize}[leftmargin=*]
    \item \texttt{Quiet(x)}: Is \textit{x} quiet?
    \item \texttt{Furry(x)}: Is \textit{x} furry?
    \item (... More predicates ...)
\end{itemize}

\textbf{Facts:}
\begin{itemize}[leftmargin=*]
    \item \texttt{Quite(Anne, True)}: Anne is quiet.
    \item (... More facts ...)
\end{itemize}

\textbf{Rules:}
\begin{itemize}[leftmargin=*]
    \item \texttt{Young(\$x, True) $\Rightarrow$ Furry(\$x, True)}: Young people are furry.
    \item (... More rules ...)
\end{itemize}

\textbf{Conditional Rules:}
\begin{itemize}
    \item \texttt{Quite(Anne, True) $\Rightarrow$ Red(\$x, True)}: If Anne is quiet then Anne is red.
    \item (... More conditional rules ...)
\end{itemize}

\textbf{Rules with Compound Predicates}
\begin{itemize}
    \item \texttt{Young(\$x, True) $\lor$ Green(\$x, True)}: Young, green people are rough.
    \item (... More conditional rules ...)
\end{itemize}

\textbf{Query:}
\begin{itemize}[leftmargin=*]
    \item White(Anne, True) ::: Anne is white.
\end{itemize}

}
\end{tcolorbox}

\begin{tcolorbox}[breakable, title=Step-2: Deriving plan based on the natural and symbolic context,colback=gray!5,colframe=nmgray!75!black,before skip=1pt, after skip=1pt,fontupper=\linespread{0.8}\selectfont]
{\footnotesize

\textbf{Task:}

Can you derive a step-by-step plan that can use the premises and First-Order Logic inference rules to infer new knowledge if needed, in order to prove the statement? Start with identifying the goal and break down the necessary logical inference step by step.

\textbf{First-Order Logic Inference Rules:}

The First-Order Logic inference rules include but are not limited to: Modus Ponens, Modus Tollens, (... More FOL rules ...).

\textbf{Predicates:}
\begin{itemize}[leftmargin=*]
    \item \texttt{Quiet(x)}: Is \textit{x} quiet?
    \item \texttt{Furry(x)}: Is \textit{x} furry?
    \item (... More predicates ...)
\end{itemize}

\textbf{Facts:}
\begin{itemize}[leftmargin=*]
    \item \texttt{Quite(Anne, True)}: Anne is quiet.
    \item (... More facts ...)
\end{itemize}

\textbf{Rules:}
\begin{itemize}[leftmargin=*]
    \item \texttt{Young(\$x, True) $\Rightarrow$ Furry(\$x, True)}: Young people are furry.
    \item (... More rules ...)
\end{itemize}

\textbf{Conditional Rules:}
\begin{itemize}
    \item \texttt{Quite(Anne, True) $\Rightarrow$ Red(\$x, True)}: If Anne is quiet then Anne is red.
    \item (... More conditional rules ...)
\end{itemize}

\textbf{Rules with Compound Predicates}
\begin{itemize}
    \item \texttt{Young(\$x, True) $\lor$ Green(\$x, True)}: Young, green people are rough.
    \item (... More conditional rules ...)
\end{itemize}

\textbf{Query:}
\begin{itemize}[leftmargin=*]
    \item White(Anne, True) ::: Anne is white.
\end{itemize}

\textbf{Plan:}

Plan to determine the truth value of "White(Anne, True)":
\begin{itemize}[leftmargin=*]
    \item 1. Identify the Goal: Our goal is to establish whether White(Anne, True)
    \item 2. Define relevant predicates and facts: We need to consider predicates related to being quiet, red, green, white and condition related to youth and roughness since these are directly or indirectly related to the query about Anne being whte.

    \item (... More plan ...)

    \item This plan systematically breaks down the steps necessary to assess the truth value of the statement concerning Anne's color based on the predicates, facts, and rules provided, applying First-Order Logic inference rules throughout the process.
\end{itemize}

}

\end{tcolorbox}

\begin{tcolorbox}[breakable, title=Step-3: Solving the Problem given the context and plan,colback=gray!5,colframe=nmgray!75!black,before skip=1pt, after skip=1pt,fontupper=\linespread{0.8}\selectfont]
{\footnotesize

\textbf{Task:}

Based on the premises, question and plan, execute each step by selecting relevant premises and making inferences based on First-Order Logic inference rules to solve the question. Ensure to follow the plan and indicate what First-Order Logic inference rules you used.

\textbf{First-Order Logic Inference Rules:}

The First-Order Logic inference rules include but are not limited to: Modus Ponens, Modus Tollens, (... More FOL rules ...)

\textbf{Predicates:}
\begin{itemize}[leftmargin=*]
    \item \texttt{Quiet(x)}: Is \textit{x} quiet?
    \item \texttt{Furry(x)}: Is \textit{x} furry?
    \item (... More predicates ...)
\end{itemize}

\textbf{Facts:}
\begin{itemize}[leftmargin=*]
    \item \texttt{Quite(Anne, True)}: Anne is quiet.
    \item (... More facts ...)
\end{itemize}

\textbf{Rules:}
\begin{itemize}[leftmargin=*]
    \item \texttt{Young(\$x, True) $\Rightarrow$ Furry(\$x, True)}: Young people are furry.
    \item (... More rules ...)
\end{itemize}

\textbf{Conditional Rules:}
\begin{itemize}
    \item \texttt{Quite(Anne, True) $\Rightarrow$ Red(\$x, True)}: If Anne is quiet then Anne is red.
    \item (... More conditional rules ...)
\end{itemize}

\textbf{Rules with Compound Predicates}
\begin{itemize}
    \item \texttt{Young(\$x, True) $\lor$ Green(\$x, True)}: Young, green people are rough.
    \item (... More conditional rules ...)
\end{itemize}

\textbf{Query:}
\begin{itemize}[leftmargin=*]
    \item White(Anne, True) ::: Anne is white.
\end{itemize}

\textbf{Plan:}

Plan to determine the truth value of "White(Anne, True)":
\begin{itemize}[leftmargin=*]
    \item 1. Identify the Goal: Our goal is to establish whether White(Anne, True)
    \item 2. Define relevant predicates and facts: We need to consider predicates related to being quiet, red, green, white and condition related to youth and roughness since these are directly or indirectly related to the query about Anne being whte.

    \item (... More plan ...)

    \item This plan systematically breaks down the steps necessary to assess the truth value of the statement concerning Anne's color based on the predicates, facts, and rules provided, applying First-Order Logic inference rules throughout the process.
\end{itemize}

\textbf{Execution:}

Let's execute the plan step by step, applying First-Order Logic inference rules.

\begin{itemize}
    \item Identify the Goal: Goal: Determine if \text{White(Anne, True)} \text{ is true.}

    \item Review given facts about Anne: \\
        - \text{Fact: } \text{Quite(Anne, True)}. \\
       - \text{Inference Rule: Given Fact.}  

    \item Analyze conditional rules involving Anne:
      \texttt{Quite(Anne, True) $\Rightarrow$ Red(\$x, True)} \\
      \text Inference Rule: Modus Ponens (Given Anne is quiet, we can infer Anne is red.)

    \item (... More solving steps ...)

    \item Given the information and the rules, we cannot directly infer "White(Anne, True)" based on the First-Order Logic inference rules provided. Thus, White(Anne, True) is unknown.
\end{itemize}
}
\end{tcolorbox}

\begin{tcolorbox}[breakable, title=Step-4: Verifying the translation and solving process ,colback=gray!5,colframe=nmgray!75!black,before skip=1pt, after skip=1pt,fontupper=\linespread{0.8}\selectfont]
{\footnotesize

\textbf{Task:}

The task is to verify whether the original execution correctly determines the value of the given conclusion based on the context.
You should abide by the First-Order Logic rules when checking the original execution and the conclusion. \\
When verifying the logical process and the final answer, make sure you fully consider the following aspects before verification: \\
1. Whether the logical process abides by the First-Order Logic inference rules. \\
2. Assumption and Consistency Verification. Please check whether the relevant facts or rules used in a logical step are indeed from the context or inferred from the previous steps. If you think it contains assumption in the logical steps, you need to clearly what assumption it is and verify whether the assumptions are from the context and whether it is already inferred from the previous steps. Please note that an assumption that does not exist from the context or was not inferred from the previous step will be considered an invalid assumption.
Please note that if an assumption is semantically equivalent to context or information inferred from the context, it should be seen as a valid assumption. For example, "X sees X" can be interpreted as "X sees itself" since they are semantically equivalent. \\
3. Whether the translation of First-Order Logic is semantically consistent with the natural language. Make sure that for compound predicates that are separated by a comma, you should treat the comma as "or". For instance, "All red, rough things are big." should be interpreted as Both red things and rough things are big. Thus, it should be translated to translated to "Red(\$x, True) $\lor$ Rough(\$x, True) $\rightarrow$ Big(\$x, True)" because being red or rough will make the conditional statement being true. \\
4. Refinement of Logical Process: If the original execution violates any of the above steps, refine the logical process. Use the relevant premises and information correctly derived from previous steps to create a more accurate logical deduction. \\
5. Handling of Conditional Statements in Conclusions: For conclusions containing conditional statements (e.g., "If..."), remember that in First-Order Logic, a conditional statement is considered false only when the antecedent is true and the consequent is false. In all other cases, the conditional statement is considered true. \\
6. If applicable, please provide a detailed analysis of each step in the logical process and a refined answer at the end, using the format {true/false/unknown}. The answer should be one of these three options.
7. Make sure the conclusion aligns with the information inferred from the logical steps.

\textbf{Predicates:}
\begin{itemize}[leftmargin=*]
    \item \texttt{Quiet(x)}: Is \textit{x} quiet?
    \item \texttt{Furry(x)}: Is \textit{x} furry?
    \item (... More predicates ...)
\end{itemize}

\textbf{Facts:}
\begin{itemize}[leftmargin=*]
    \item \texttt{Quite(Anne, True)}: Anne is quiet.
    \item (... More facts ...)
\end{itemize}

\textbf{Rules:}
\begin{itemize}[leftmargin=*]
    \item \texttt{Young(\$x, True) $\Rightarrow$ Furry(\$x, True)}: Young people are furry.
    \item (... More rules ...)
\end{itemize}

\textbf{Conditional Rules:}
\begin{itemize}
    \item \texttt{Quite(Anne, True) $\Rightarrow$ Red(\$x, True)}: If Anne is quiet then Anne is red.
    \item (... More conditional rules ...)
\end{itemize}

\textbf{Rules with Compound Predicates}
\begin{itemize}
    \item \texttt{Young(\$x, True) $\lor$ Green(\$x, True)}: Young, green people are rough.
    \item (... More conditional rules ...)
\end{itemize}

\textbf{Query:}
\begin{itemize}[leftmargin=*]
    \item White(Anne, True) ::: Anne is white.
\end{itemize}

\textbf{Original Execution:}

Let's execute the plan step by step, applying First-Order Logic inference rules.

\begin{itemize}
    \item Identify the Goal: Goal: Determine if \text{White(Anne, True)} \text{ is true.}

    \item Review given facts about Anne: \\
        - \text{Fact: } \text{Quite(Anne, True)}. \\
       - \text{Inference Rule: Given Fact.}  

    \item Analyze conditional rules involving Anne:
      \texttt{Quite(Anne, True) $\Rightarrow$ Red(\$x, True)} \\
      \text Inference Rule: Modus Ponens (Given Anne is quiet, we can infer Anne is red.)

    \item (... More solving steps ...)

    \item Given the information and the rules, we cannot directly infer "White(Anne, True)" based on the First-Order Logic inference rules provided. Thus, White(Anne, True) is unknown.
\end{itemize}

\textbf{Verification:}

To verify the logical validity of the original execution concerning the statement "White(Anne, True)" based on the given premises, let's assess each step against the requirements of First-Order Logic, assumptions, consistency, and semantic equivalence.\\
\textbf{Context verification:}
\begin{itemize}
    
    \item  The context and query correctly correspond to the information in the original context and are semantically consistent.

\end{itemize}

\textbf{Logical Process Verification:}
\begin{itemize}
    \item 1. Fact: Quite(Anne, True) is directly from the context with no invalid assumption.
    \item 2. The rule "Quite(Anne, True) $\rightarrow$ Red(\$x, True)" also contains a typo and should be "Quiet(Anne, True) $\rightarrow$ Red(\$x, True)." This rule incorrectly uses \$x instead of specifically stating "Anne," which might lead to confusion. The correct interpretation for the context should directly indicate that if Anne is quiet, then Anne is red, implying "Quiet(Anne, True) $\rightarrow$ Red(Anne, True)."
    \item (... More logical verification ...)
    \item The conclusion that "White(Anne, True)" remains unknown is consistent with the logical assessment of the provided context, considering the application of First-Order Logic inference rules and the available premises. 
\end{itemize}

}
\end{tcolorbox}

\subsubsection{FOLIO}

\begin{tcolorbox}[breakable, title=Step-1: Translating natural language context into symbolic,colback=gray!5,colframe=nmgray!75!black,before skip=1pt, after skip=1pt,fontupper=\linespread{0.8}\selectfont]
{\footnotesize

\textbf{Task Description:}

Given a problem description and a question, the task is to parse the problem and the question into First-Order Logic formulas. The grammar of the First-Order Logic formula is defined as follows:
\begin{enumerate}
    \item Logical conjunction of \texttt{expr1} and \texttt{expr2}: \( \texttt{expr1} \land \texttt{expr2} \)
    \item Logical disjunction of \texttt{expr1} and \texttt{expr2}: \( \texttt{expr1} \lor \texttt{expr2} \)
    \item Logical exclusive disjunction of \texttt{expr1} and \texttt{expr2}: \( \texttt{expr1} \oplus \texttt{expr2} \)
    \item Logical negation of \texttt{expr1}: \( \lnot\texttt{expr1} \)
    \item \texttt{expr1} implies \texttt{expr2}: \( \texttt{expr1} \rightarrow \texttt{expr2} \)
    \item \texttt{expr1} if and only if \texttt{expr2}: \( \texttt{expr1} \leftrightarrow \texttt{expr2} \)
    \item Logical universal quantification: \( \forall x \)
    \item Logical existential quantification: \( \exists x \)
\end{enumerate}

\textbf{Problem:}
\begin{itemize}[leftmargin=*]
    \item Miroslav Venhoda was a Czech choral conductor who specialized in the performance of Renaissance and Baroque music. Any choral conductor is a musician. Some musicians love music. Miroslav Venhoda published a book in 1946 called Method of Studying Gregorian Chant.
\end{itemize}

\textbf{Question:}
\begin{itemize}[leftmargin=*]
    \item Is the following statement true, false, or uncertain? Miroslav Venhoda loved music.
\end{itemize}

\textbf{Translation: }\\
\textbf{Predicates:}
\begin{itemize}
    \item $\text{Czech}(x)$: $x$ is a Czech person.
    \item $\text{ChoralConductor}(x)$: $x$ is a choral conductor.
    \item $\text{Musician}(x)$: $x$ is a musician.
    \item $\text{Love}(x, y)$: $x$ loves $y$.
    \item $\text{Author}(x, y)$: $x$ is the author of $y$.
    \item $\text{Book}(x)$: $x$ is a book.
    \item $\text{Publish}(x, y)$: $x$ is published in year $y$.
    \item $\text{Specialize}(x, y)$: $x$ specializes in $y$.
\end{itemize}

\textbf{Premises:}
\begin{itemize}
    \item $\text{Czech}(\text{miroslav}) \land \text{ChoralConductor}(\text{miroslav}) \land \text{Specialize}(\text{miroslav}, \text{renaissance}) \land \text{Specialize}(\text{miroslav}, \text{baroque})$: Miroslav Venhoda was a Czech choral conductor who specialized in the performance of Renaissance and Baroque music.
    \item $\forall x (\text{ChoralConductor}(x) \rightarrow \text{Musician}(x))$: Any choral conductor is a musician.
    \item $\exists x (\text{Musician}(x) \land \text{Love}(x, \text{music}))$: Some musicians love music.
    \item $\text{Book}(\text{methodOfStudyingGregorianChant}) \land \text{Author}(\text{miroslav}, \text{methodOfStudyingGregorianChant}) \land \text{Publish}(\text{methodOfStudyingGregorianChant}, \text{year1946})$: Miroslav Venhoda published a book in 1946 called Method of Studying Gregorian Chant.
\end{itemize}

\textbf{Query:}
\begin{itemize}[leftmargin=*]
    \item Love(miroslav, music) ::: Miroslav Venhoda loved music.
\end{itemize}

}
\end{tcolorbox}

\begin{tcolorbox}[breakable, title=Step-2: Deriving plan based on the natural and symbolic context,colback=gray!5,colframe=nmgray!75!black,before skip=1pt, after skip=1pt,fontupper=\linespread{0.8}\selectfont]
{\footnotesize

\textbf{Task:}

Can you derive a step-by-step plan that can use the premises and First-Order Logic inference rules to infer new knowledge if needed, in order to prove the statement? Start with identifying the goal and break down the necessary logical inference step by step.

\textbf{First-Order Logic Inference Rules:}

The First-Order Logic inference rules include but are not limited to: Modus Ponens, Modus Tollens, (... More FOL rules ...).

\textbf{Predicates:}
\begin{itemize}
    \item $\text{Czech}(x)$: $x$ is a Czech person.
    \item $\text{ChoralConductor}(x)$: $x$ is a choral conductor.
    \item $\text{Musician}(x)$: $x$ is a musician.
    \item $\text{Love}(x, y)$: $x$ loves $y$.
    \item $\text{Author}(x, y)$: $x$ is the author of $y$.
    \item $\text{Book}(x)$: $x$ is a book.
    \item $\text{Publish}(x, y)$: $x$ is published in year $y$.
    \item $\text{Specialize}(x, y)$: $x$ specializes in $y$.
\end{itemize}

\textbf{Premises:}
\begin{itemize}
    \item $\text{Czech}(\text{miroslav}) \land \text{ChoralConductor}(\text{miroslav}) \land \text{Specialize}(\text{miroslav}, \text{renaissance}) \land \text{Specialize}(\text{miroslav}, \text{baroque})$: Miroslav Venhoda was a Czech choral conductor who specialized in the performance of Renaissance and Baroque music.
    \item $\forall x (\text{ChoralConductor}(x) \rightarrow \text{Musician}(x))$: Any choral conductor is a musician.
    \item $\exists x (\text{Musician}(x) \land \text{Love}(x, \text{music}))$: Some musicians love music.
    \item $\text{Book}(\text{methodOfStudyingGregorianChant}) \land \text{Author}(\text{miroslav}, \text{methodOfStudyingGregorianChant}) \land \text{Publish}(\text{methodOfStudyingGregorianChant}, \text{year1946})$: Miroslav Venhoda published a book in 1946 called Method of Studying Gregorian Chant.
\end{itemize}

\textbf{Query:}
\begin{itemize}[leftmargin=*]
    \item Love(miroslav, music) ::: Miroslav Venhoda loved music.
\end{itemize}

\textbf{Plan:}

To determine whether the conclusion "Love(miroslav, music) ::: Miroslav Venhoda loved music" is true, false, or unknown based on the given premises, we can follow these steps:
\begin{itemize}[leftmargin=*]
    \item 1. Identify the Goal: Our goal is to assess whether it can be logically concluded that Miroslav Venhoda loved music based on the available premises.
    \item 2. Define relevant predicates and facts: Premise 1 introduces Miroslav as a Czech choral conductor specializing in Renaissance and Baroque music. This establishes his professional background and interests but does not directly address his personal feelings toward music.
    Premise 2 establishes that all choral conductors are musicians. This is a classification premise that links the profession of being a choral conductor with being a musician.

    \item (... More plan ...)

    \item This plan outlines a systematic approach to assessing the provided conclusion against the premises using First-Order Logic inference rules.
\end{itemize}

}

\end{tcolorbox}

\begin{tcolorbox}[breakable, title=Step-3: Solving the Problem given the context and plan,colback=gray!5,colframe=nmgray!75!black,before skip=1pt, after skip=1pt,fontupper=\linespread{0.8}\selectfont]
{\footnotesize

\textbf{Task:}

Based on the premises, question, and plan, execute each step by selecting relevant premises and making inferences based on First-Order Logic inference rules to solve the question. Ensure to follow the plan and indicate what First-Order Logic inference rules you used.

\textbf{First-Order Logic Inference Rules:}

The First-Order Logic inference rules include but are not limited to: Modus Ponens, Modus Tollens, (... More FOL rules ...)

\textbf{Predicates:}
\begin{itemize}
    \item $\text{Czech}(x)$: $x$ is a Czech person.
    \item $\text{ChoralConductor}(x)$: $x$ is a choral conductor.
    \item $\text{Musician}(x)$: $x$ is a musician.
    \item $\text{Love}(x, y)$: $x$ loves $y$.
    \item $\text{Author}(x, y)$: $x$ is the author of $y$.
    \item $\text{Book}(x)$: $x$ is a book.
    \item $\text{Publish}(x, y)$: $x$ is published in year $y$.
    \item $\text{Specialize}(x, y)$: $x$ specializes in $y$.
\end{itemize}

\textbf{Premises:}
\begin{itemize}
    \item $\text{Czech}(\text{miroslav}) \land \text{ChoralConductor}(\text{miroslav}) \land \text{Specialize}(\text{miroslav}, \text{renaissance}) \land \text{Specialize}(\text{miroslav}, \text{baroque})$: Miroslav Venhoda was a Czech choral conductor who specialized in the performance of Renaissance and Baroque music.
    \item $\forall x (\text{ChoralConductor}(x) \rightarrow \text{Musician}(x))$: Any choral conductor is a musician.
    \item $\exists x (\text{Musician}(x) \land \text{Love}(x, \text{music}))$: Some musicians love music.
    \item $\text{Book}(\text{methodOfStudyingGregorianChant}) \land \text{Author}(\text{miroslav}, \text{methodOfStudyingGregorianChant}) \land \text{Publish}(\text{methodOfStudyingGregorianChant}, \text{year1946})$: Miroslav Venhoda published a book in 1946 called Method of Studying Gregorian Chant.
\end{itemize}

\textbf{Query:}
\begin{itemize}[leftmargin=*]
    \item Love(miroslav, music) ::: Miroslav Venhoda loved music.
\end{itemize}

\textbf{Plan:}

To determine whether the conclusion "Love(miroslav, music) ::: Miroslav Venhoda loved music" is true, false, or unknown based on the given premises, we can follow these steps:
\begin{itemize}[leftmargin=*]
    \item 1. Identify the Goal: Our goal is to assess whether it can be logically concluded that Miroslav Venhoda loved music based on the available premises.
    \item 2. Define relevant predicates and facts: Premise 1 introduces Miroslav as a Czech choral conductor specializing in Renaissance and Baroque music. This establishes his professional background and interests but does not directly address his personal feelings toward music.
    Premise 2 establishes that all choral conductors are musicians. This is a classification premise that links the profession of being a choral conductor with being a musician.

    \item (... More plan ...)

    \item This plan outlines a systematic approach to assessing the provided conclusion against the premises using First-Order Logic inference rules.
\end{itemize}

\textbf{Execution:}

To solve the question about whether "Miroslav Venhoda loved music" based on the given premises, let's proceed systematically:

\begin{itemize}
    \item Identify the Goal: Goal: Determine if \text{Love(miroslav, music)} \text{ is true.}

    \item Apply Universal Instantiation to Premise 2: From Premise 2, since Miroslav is a choral conductor, we can infer he is also a musician based on the rule that all choral conductors are musicians. Using Universal Instantiation and Modus Ponens, we deduce Musician(miroslav).

    \item (... More solving steps ...)

    \item The conclusion that "Miroslav Venhoda loved music" cannot be definitively proven true solely based on the provided premises and the strict application of First-Order Logic rules. Therefore, the answer is unknown.
\end{itemize}
}
\end{tcolorbox}

\begin{tcolorbox}[breakable, title=Step-4: Verifying the translation and solving process ,colback=gray!5,colframe=nmgray!75!black,before skip=1pt, after skip=1pt,fontupper=\linespread{0.8}\selectfont]
{\footnotesize

\textbf{Task:}

The task is to verify whether the original execution correctly determines the value of the given conclusion based on the context.
You should abide by the First-Order Logic rules when checking the original execution and the conclusion. \\
When verifying the logical process and the final answer, make sure you fully consider the following aspects before verification: \\
1. Whether the logical process abides by the First-Order Logic inference rules. \\
2. Assumption and Consistency Verification. Please check whether the relevant facts or rules used in a logical step are indeed from the context or inferred from the previous steps. If you think it contains assumption in the logical steps, you need to clearly what assumption it is and verify whether the assumptions are from the context and whether it is already inferred from the previous steps. Please note that an assumption that does not exist from the context or was not inferred from the previous step will be considered an invalid assumption.
Please note that if an assumption is semantically equivalent to context or information inferred from the context, it should be seen as a valid assumption. For example, "X sees X" can be interpreted as "X sees itself" since they are semantically equivalent. \\
3. Whether the translation First-Order Logic is semantically consistent with the natural language. Make sure that for compound predicates that are separated by a comma, you should treat the comma as "or". For instance, "All red, rough things are big." should be interpreted as Both red things and rough things are big. Thus, it should be translated to translated to "Red(\$x, True) $\lor$ Rough(\$x, True) $\rightarrow$ Big(\$x, True)" because being red or rough will make the conditional statement being true. \\
4. Refinement of Logical Process: If the original execution violates any of the above steps, refine the logical process. Use the relevant premises and information correctly derived from previous steps to create a more accurate logical deduction. \\
5. Handling of Conditional Statements in Conclusions: For conclusions containing conditional statements (e.g., "If..."), remember that in First-Order Logic, a conditional statement is considered false only when the antecedent is true and the consequent is false. In all other cases, the conditional statement is considered true. \\
6. If applicable, please provide a detailed analysis of each step in the logical process and a refined answer at the end, using the format {true/false/unknown}. The answer should be one of these three options.
7. Make sure the conclusion aligns with the information inferred from the logical steps.

\textbf{Predicates:}
\begin{itemize}
    \item $\text{Czech}(x)$: $x$ is a Czech person.
    \item $\text{ChoralConductor}(x)$: $x$ is a choral conductor.
    \item $\text{Musician}(x)$: $x$ is a musician.
    \item $\text{Love}(x, y)$: $x$ loves $y$.
    \item $\text{Author}(x, y)$: $x$ is the author of $y$.
    \item $\text{Book}(x)$: $x$ is a book.
    \item $\text{Publish}(x, y)$: $x$ is published in year $y$.
    \item $\text{Specialize}(x, y)$: $x$ specializes in $y$.
\end{itemize}

\textbf{Premises:}
\begin{itemize}
    \item $\text{Czech}(\text{miroslav}) \land \text{ChoralConductor}(\text{miroslav}) \land \text{Specialize}(\text{miroslav}, \text{renaissance}) \land \text{Specialize}(\text{miroslav}, \text{baroque})$: Miroslav Venhoda was a Czech choral conductor who specialized in the performance of Renaissance and Baroque music.
    \item $\forall x (\text{ChoralConductor}(x) \rightarrow \text{Musician}(x))$: Any choral conductor is a musician.
    \item $\exists x (\text{Musician}(x) \land \text{Love}(x, \text{music}))$: Some musicians love music.
    \item $\text{Book}(\text{methodOfStudyingGregorianChant}) \land \text{Author}(\text{miroslav}, \text{methodOfStudyingGregorianChant}) \land \text{Publish}(\text{methodOfStudyingGregorianChant}, \text{year1946})$: Miroslav Venhoda published a book in 1946 called Method of Studying Gregorian Chant.
\end{itemize}

\textbf{Query:}
\begin{itemize}[leftmargin=*]
    \item Love(miroslav, music) ::: Miroslav Venhoda loved music.
\end{itemize}

\textbf{Original Execution:}

To solve the question about whether "Miroslav Venhoda loved music" based on the given premises, let's proceed systematically:

\begin{itemize}
    \item Identify the Goal: Goal: Determine if \text{Love(miroslav, music)} \text{ is true.}

    \item Apply Universal Instantiation to Premise 2: From Premise 2, since Miroslav is a choral conductor, we can infer he is also a musician based on the rule that all choral conductors are musicians. Using Universal Instantiation and Modus Ponens, we deduce Musician(miroslav).

    \item (... More solving steps ...)

    \item The conclusion that "Miroslav Venhoda loved music" cannot be definitively proven true solely based on the provided premises and the strict application of First-Order Logic rules. Therefore, the answer is unknown.
\end{itemize}

\textbf{Verification:}

Verification of the original execution involves evaluating the logical process used to determine whether Miroslav Venhoda loved music, based on the provided context and premises. \\
\textbf{Context verification:}
\begin{itemize}
    
    \item  The context and query correctly correspond to the information in the original context and are semantically consistent.

\end{itemize}

\textbf{Logical Process Verification:}
\begin{itemize}
    \item 1. Correctly identify the goal.
    \item 2. Correctly use Universal Instantiation and Modus Ponens to infer Musician(miroslav).
    \item (... More logical verification ...)
    \item The original execution's conclusion that the statement "Love(miroslav, music)" is unknown is correct based on the available premises. The verification process confirms that the original logical process and conclusion are accurate, adhering to the principles of First-Order Logic given the information provided.
\end{itemize}

}
\end{tcolorbox}

\subsubsection{LogicalDeduction}

\begin{tcolorbox}[breakable, title=Step-1: Translating natural language context into symbolic,colback=gray!5,colframe=nmgray!75!black,before skip=1pt, after skip=1pt,fontupper=\linespread{0.8}\selectfont]
{\footnotesize

\textbf{Task Description:}

Task Description: You are given a problem description. The task is to parse the problem as a constraint satisfaction problem, defining the domain, variables, and constraints.

\textbf{Problem:}
\begin{itemize}[leftmargin=*]
    \item The following paragraphs each describe a set of three objects arranged in a fixed order. The statements are logically consistent within each paragraph. In an antique car show, there are three vehicles: a station wagon, a convertible, and a minivan. The station wagon is the oldest. The minivan is newer than the convertible.
\end{itemize}

\textbf{Question:}
\begin{itemize}[leftmargin=*]
    \item Which of the following is true?
\end{itemize}

\textbf{Options:}
\begin{itemize}[leftmargin=*]
    \item A) The station wagon is the second-newest.
    \item B) The convertible is the second-newest.
    \item C) The minivan is the second-newest.
\end{itemize}

\textbf{Translation: }\\
\textbf{Domain:}
\begin{itemize}
    \item 1: oldest
    \item 3: newest
\end{itemize}

\textbf{Variables:}
\begin{itemize}
    \item station\_wagon \(\in \{1, 2, 3\}\)
    \item convertible \(\in \{1, 2, 3\}\)
    \item minivan \(\in \{1, 2, 3\}\)
\end{itemize}

\textbf{Constraints:}
\begin{itemize}
    \item \( \text{station\_wagon} == 1 \): The station wagon is the oldest.
    \item \( \text{minivan} > \text{convertible} \): The minivan is newer than the convertible.
    \item AllDifferentConstraint(\([ \text{station\_wagon}, \text{convertible}, \\ \text{minivan} ]\)): All vehicles have different values.
\end{itemize}

\textbf{Query:}
\begin{itemize}
    \item \(A) \text{station\_wagon} == 2\): The station wagon is the second-newest.
    \item \(B) \text{convertible} == 2\): The convertible is the second-newest.
    \item \(C) \text{minivan} == 2\): The minivan is the second-newest.
\end{itemize}

}
\end{tcolorbox}

\begin{tcolorbox}[breakable, title=Step-2: Deriving plan based on the natural and symbolic context,colback=gray!5,colframe=nmgray!75!black,before skip=1pt, after skip=1pt,fontupper=\linespread{0.8}\selectfont]
{\footnotesize

\textbf{Task:}

Can you derive a step-by-step plan that can use the domain, variables and constraints relevant inference rules to choose the correct option satisfying all the constraints?

\textbf{Domain:}
\begin{itemize}
    \item 1: oldest
    \item 3: newest
\end{itemize}

\textbf{Variables:}
\begin{itemize}
    \item station\_wagon \(\in \{1, 2, 3\}\)
    \item convertible \(\in \{1, 2, 3\}\)
    \item minivan \(\in \{1, 2, 3\}\)
\end{itemize}

\textbf{Constraints:}
\begin{itemize}
    \item \( \text{station\_wagon} == 1 \): The station wagon is the oldest.
    \item \( \text{minivan} > \text{convertible} \): The minivan is newer than the convertible.
    \item AllDifferentConstraint(\([ \text{station\_wagon}, \text{convertible}, \\ \text{minivan} ]\)): All vehicles have different values.
\end{itemize}

\textbf{Query:}
\begin{itemize}
    \item \(A) \text{station\_wagon} == 2\): The station wagon is the second-newest.
    \item \(B) \text{convertible} == 2\): The convertible is the second-newest.
    \item \(C) \text{minivan} == 2\): The minivan is the second-newest.
\end{itemize}

\textbf{Plan:}
To approach the constraint optimization problem described, involving variables like station\_wagon, convertible, and minivan with their respective constraints, you can follow a systematic plan.

\begin{itemize}[leftmargin=*]
    \item 1. Identify Variables and their domain: List all the variables (station\_wagon, convertible, minivan) and note their possible values (1, 2, 3), where 1 represents the oldest and 3 represents the newest.
    \item 2. Apply the constraints to narrow down the possibility: The station\_wagon is the oldest.  \\
    The minivan is newer than the convertible. \\
    All vehicles (station\_wagon, convertible, minivan) must have different ages.

    \item (... More plan ...)

    \item This plan provides a systematic approach to solving the problem without directly computing the solution. It emphasizes understanding constraints, logical deduction, and iterative refinement to navigate toward a solution that satisfies all given conditions.
\end{itemize}

}

\end{tcolorbox}

\begin{tcolorbox}[breakable, title=Step-3: Solving the Problem given the context and plan,colback=gray!5,colframe=nmgray!75!black,before skip=1pt, after skip=1pt,fontupper=\linespread{0.8}\selectfont]
{\footnotesize

\textbf{Task:}

The task is a constraint optimization problem, and you need to select the most appropriate option that is true from the given options.
The domain, variables, constraints, and relevant plan will be given to you to solve the question. \\

There will be only one answer, so choose the one you think is the most likely. \\

Here is a workflow you need to follow to solve the constraint optimization problem: \\
1.Domain and Variables: Identify the possible values (domain) and variables involved. \\
2.Apply Constraints: Read and apply the specific constraints to the variables. \\
3.List Possibilities: Enumerate all potential combinations or orders that could satisfy the constraints. \\
4.Determine Order/Values: Establish the correct order or specific values for each variable as per the constraints. \\
5.Demonstrate Satisfaction of Constraints: Provide a specific example that shows all constraints are met. \\
6.Evaluate Queries: Analyze each query against the established order or values. \\
7.Identify Correct Answer: Choose the option that aligns with the determined order/values. \\
8.Present Final Answer: Clearly state the solution that satisfies the constraints and query. \\

\textbf{Domain:}
\begin{itemize}
    \item 1: oldest
    \item 3: newest
\end{itemize}

\textbf{Variables:}
\begin{itemize}
    \item station\_wagon \(\in \{1, 2, 3\}\)
    \item convertible \(\in \{1, 2, 3\}\)
    \item minivan \(\in \{1, 2, 3\}\)
\end{itemize}

\textbf{Constraints:}
\begin{itemize}
    \item \( \text{station\_wagon} == 1 \): The station wagon is the oldest.
    \item \( \text{minivan} > \text{convertible} \): The minivan is newer than the convertible.
    \item AllDifferentConstraint(\([ \text{station\_wagon}, \text{convertible}, \\ \text{minivan} ]\)): All vehicles have different values.
\end{itemize}

\textbf{Query:}
\begin{itemize}
    \item \(A) \text{station\_wagon} == 2\): The station wagon is the second-newest.
    \item \(B) \text{convertible} == 2\): The convertible is the second-newest.
    \item \(C) \text{minivan} == 2\): The minivan is the second-newest.
\end{itemize}

\textbf{Plan:}
To approach the constraint optimization problem described, involving variables like station\_wagon, convertible, and minivan with their respective constraints, you can follow a systematic plan.

\begin{itemize}[leftmargin=*]
    \item 1. Identify Variables and their domain: List all the variables (station\_wagon, convertible, minivan) and note their possible values (1, 2, 3), where 1 represents the oldest and 3 represents the newest.
    \item 2. Apply the constraints to narrow down the possibility: The station\_wagon is the oldest.  \\
    The minivan is newer than the convertible. \\
    All vehicles (station\_wagon, convertible, minivan) must have different ages.

    \item (... More plan ...)

    \item This plan provides a systematic approach to solving the problem without directly computing the solution. It emphasizes understanding constraints, logical deduction, and iterative refinement to navigate toward a solution that satisfies all given conditions.
\end{itemize}

\textbf{Execution:}

Let's solve the given constraint optimization problem following the outlined plan:

\begin{itemize}
    \item Identify variables and their domains: Variables: station\_wagon, convertible, minivan. Domain for each: 1 (oldest), 2, 3 (newest)

    \item station\_wagon == 1: Station wagon is the oldest. \\ minivan > convertible: Minivan is newer than the convertible. \\ All vehicles have different ages.

    \item Apply Direct Constraints: Assign station\_wagon the value 1 since it's the oldest.

    \item (... More solving steps ...)

    \item Based on the configurations that satisfy all constraints, the only query that holds true is convertible == 2, making it the second-newest vehicle. Therefore, the final answer is B.
\end{itemize}
}
\end{tcolorbox}

\begin{tcolorbox}[breakable, title=Step-4: Verifying the translation and solving process ,colback=gray!5,colframe=nmgray!75!black,before skip=1pt, after skip=1pt,fontupper=\linespread{0.8}\selectfont]
{\footnotesize

\textbf{Task:}

Please verify the solution to the following constraint optimization problem using a detailed method. The problem consists of a set of variables, each with a specific domain, and a set of constraints. Your task is to determine if the proposed solution in the "Original execution" correctly satisfies all the constraints, and if not, find a valid solution that does. \\
The context will contain the following: \\
- Domain: Specify the domain for the problem, for example, '1: highest to 5: lowest' or '1: oldest to 5: newest'. \\
- Variables: List all the variables in the problem with their possible values in the domain. \\
- Constraints: Describe each constraint in detail, making sure to interpret them correctly according to the domain. For example, 'X > Y' means 'X finishes below Y' if the domain is '1: highest to 5: lowest'. \\
- Query: State the query that needs to be answered based on the variables and constraints. \\ 
Solution Verification Steps: \\
1. Review the domain, variables, constraints, and query between symbolic format and natural language for consistency with explanation. If there is an inconsistency between the symbolic format and the natural language description of constraints, make sure you only change the symbolic format but DON'T change the natural language as natural language is the ground truth standard. \\
2. If the proposed solution violates any constraints, attempt to find a valid order that satisfies all constraints. Explain your reasoning for the new order. \\
3. Once a valid order is found, evaluate it against the query. \\
4. Present the final answer based on your evaluation. \\
Your task is to apply this method to verify the problem, and if necessary, re-solve it to find a valid solution. \\
Note: Make sure to interpret the constraints accurately based on the domain, and carefully evaluate each constraint against the proposed solution. If the solution is incorrect, methodically find a valid order that satisfies all constraints, and then answer the query based on this valid order.

\textbf{Domain:}
\begin{itemize}
    \item 1: oldest
    \item 3: newest
\end{itemize}

\textbf{Variables:}
\begin{itemize}
    \item station\_wagon \(\in \{1, 2, 3\}\)
    \item convertible \(\in \{1, 2, 3\}\)
    \item minivan \(\in \{1, 2, 3\}\)
\end{itemize}

\textbf{Constraints:}
\begin{itemize}
    \item \( \text{station\_wagon} == 1 \): The station wagon is the oldest.
    \item \( \text{minivan} > \text{convertible} \): The minivan is newer than the convertible.
    \item AllDifferentConstraint(\([ \text{station\_wagon}, \text{convertible}, \\ \text{minivan} ]\)): All vehicles have different values.
\end{itemize}

\textbf{Query:}
\begin{itemize}
    \item \(A) \text{station\_wagon} == 2\): The station wagon is the second-newest.
    \item \(B) \text{convertible} == 2\): The convertible is the second-newest.
    \item \(C) \text{minivan} == 2\): The minivan is the second-newest.
\end{itemize}

\textbf{Original Execution:}

Let's solve the given constraint optimization problem following the outlined plan:

\begin{itemize}
    \item Identify variables and their domains: Variables: station\_wagon, convertible, minivan. Domain for each: 1 (oldest), 2, 3 (newest)

    \item station\_wagon == 1: Station wagon is the oldest. \\ minivan > convertible: Minivan is newer than the convertible. \\ All vehicles have different ages.

    \item Apply Direct Constraints: Assign station\_wagon the value 1 since it's the oldest.

    \item (... More solving steps ...)

    \item Based on the configurations that satisfy all constraints, the only query that holds true is convertible == 2, making it the second-newest vehicle. Therefore, the final answer is B.
\end{itemize}

\textbf{Verification:}

Verification of the translation and the Original Execution: \\
\textbf{Context verification:}
\begin{itemize}
    
    \item  The context and query correctly correspond to the information in the original context and are semantically consistent.

\end{itemize}

\textbf{Logical Process Verification:}
\begin{itemize}
    \item Given the constraints and the domain's interpretation, let's verify the original execution:
    \item 1. Station Wagon's Age: As the oldest, its correct assignment is 1. This matches both the symbolic and natural language descriptions.
    \item 2. Convertible and Minivan Relationship: The constraint "minivan > convertible" suggests the minivan is newer (higher number) than the convertible. Given the domain and the direct constraint on the station wagon, the convertible cannot be the oldest (1), and the minivan cannot be the oldest due to its relationship with the convertible. This necessitates a careful review of the possible configurations.
    \item (... More logical verification ...)
    \item The original solution's interpretation of the constraints and the final answer is consistent with the domain's direction and the natural language description of the constraints. The logic applied matches the expected outcomes, and no changes are necessary to the symbolic format or the conclusion based on the provided context and constraints. Therefore, the original answer B remains unchanged.
\end{itemize}

}
\end{tcolorbox}
\subsubsection{AR-LSAT}
.
\begin{tcolorbox}[breakable, title=Step-1: Translating natural language context into symbolic,colback=gray!5,colframe=nmgray!75!black,before skip=0pt, after skip=0pt,fontupper=\linespread{0.8}\selectfont]
{\footnotesize

\textbf{Task Description:}

Task Description: You are given a problem description. The task is to parse the problem as a constraint satisfaction problem, defining the domain, variables, and constraints.

\textbf{Problem:}
\begin{itemize}[leftmargin=*]
    \item Four boys—Fred, Juan, Marc, and Paul—and three girls—Nita, Rachel, and Trisha—will be assigned to a row of five adjacent lockers, numbered consecutively 1 through 5, arranged along a straight wall. The following conditions govern the assignment of lockers to the seven children: Each locker must be assigned to either one or two children, and each child must be assigned to exactly one locker. Each shared locker must be assigned to one girl and one boy. Juan must share a locker, but Rachel cannot share a locker. Nita's locker cannot be adjacent to Trisha's locker. Fred must be assigned to locker 3.
\end{itemize}

\textbf{Question:}
\begin{itemize}[leftmargin=*]
    \item If the first three lockers are assigned to girls, which one of the following must be true?
\end{itemize}

\textbf{Options:}
\begin{itemize}[leftmargin=*]
    \item A) Juan is assigned to locker 1.
    \item B) Nita is assigned to locker 3.
    \item C) Trisha is assigned to locker 1.
    \item D) Juan is assigned to the same locker as Trisha.
    \item E) Paul is assigned to the same locker as Trisha.
\end{itemize}

\textbf{Translation: }\\
\textbf{Domain}

\begin{itemize}
    \item Lockers: $1$ to $5$
\end{itemize}

\textbf{Variables}

\begin{itemize}
    \item Fred\_locker $\in \{1, 2, 3, 4, 5\}$
    \item Juan\_locker $\in \{1, 2, 3, 4, 5\}$
    \item Marc\_locker $\in \{1, 2, 3, 4, 5\}$
    \item Paul\_locker $\in \{1, 2, 3, 4, 5\}$
    \item Nita\_locker $\in \{1, 2, 3, 4, 5\}$
    \item Rachel\_locker $\in \{1, 2, 3, 4, 5\}$
    \item Trisha\_locker $\in \{1, 2, 3, 4, 5\}$
\end{itemize}

\textbf{Constraints}

\begin{enumerate}
    \item \textbf{Each Locker One or Two Children:} Each locker can have either one or two children assigned to it.
    
    \item \textbf{Each Child One Locker:} Each child is assigned to exactly one locker.
    
    \item \textbf{Shared Locker Boy and Girl:} If a locker is shared, it must be shared between one boy and one girl.
    
    \item \textbf{Juan Must Share:} Juan\_locker must be the same as one girl's locker.
    
    \item \textbf{Rachel Cannot Share:} Rachel\_locker must be unique and not shared with any other child.
    
    \item \textbf{Nita Not Adjacent to Trisha:} $|Nita\_locker - Trisha\_locker| \neq 1$
    
    \item \textbf{Fred Assigned to Locker 3:} Fred\_locker = 3
    
    \item \textbf{First Three Lockers to Girls:} At least one girl must be assigned to each of the lockers $1$, $2$, and $3$.
\end{enumerate}

\textbf{Queries for Options (If the first three lockers are assigned to girls, which one of the following must be true?)}

\begin{itemize}
    \item \textbf{A) Juan is assigned to locker 1.} Query: Juan\_locker = 1 and Juan\_locker is shared with a girl
    
    \item \textbf{B) Nita is assigned to locker 3.} Query: Nita\_locker = 3
    
    \item \textbf{C) Trisha is assigned to locker 1.} Query: Trisha\_locker = 1
    
    \item \textbf{D) Juan is assigned to the same locker as Trisha.} Query: Juan\_locker = Trisha\_locker
    
    \item \textbf{E) Paul is assigned to the same locker as Trisha.} Query: Paul\_locker = Trisha\_locker
\end{itemize}

}
\end{tcolorbox}

\begin{tcolorbox}[breakable, title=Step-2: Deriving plan based on the natural and symbolic context,colback=gray!5,colframe=nmgray!75!black,before skip=1pt, after skip=1pt,fontupper=\linespread{0.8}\selectfont]
{\footnotesize

\textbf{Task:}

Can you derive a step-by-step plan that can use the domain, variables and constraints relevant inference rules to choose the correct option satisfying all the constraints?

\textbf{Domain}

\begin{itemize}
    \item Lockers: $1$ to $5$
\end{itemize}

\textbf{Variables}

\begin{itemize}
    \item Fred\_locker $\in \{1, 2, 3, 4, 5\}$
    \item Juan\_locker $\in \{1, 2, 3, 4, 5\}$
    \item Marc\_locker $\in \{1, 2, 3, 4, 5\}$
    \item Paul\_locker $\in \{1, 2, 3, 4, 5\}$
    \item Nita\_locker $\in \{1, 2, 3, 4, 5\}$
    \item Rachel\_locker $\in \{1, 2, 3, 4, 5\}$
    \item Trisha\_locker $\in \{1, 2, 3, 4, 5\}$
\end{itemize}

\textbf{Constraints}

\begin{enumerate}
    \item \textbf{Each Locker One or Two Children:} Each locker can have either one or two children assigned to it.
    
    \item \textbf{Each Child One Locker:} Each child is assigned to exactly one locker.
    
    \item \textbf{Shared Locker Boy and Girl:} If a locker is shared, it must be shared between one boy and one girl.
    
    \item \textbf{Juan Must Share:} Juan\_locker must be the same as one girl's locker.
    
    \item \textbf{Rachel Cannot Share:} Rachel\_locker must be unique and not shared with any other child.
    
    \item \textbf{Nita Not Adjacent to Trisha:} $|Nita\_locker - Trisha\_locker| \neq 1$
    
    \item \textbf{Fred Assigned to Locker 3:} Fred\_locker = 3
    
    \item \textbf{First Three Lockers to Girls:} At least one girl must be assigned to each of the lockers $1$, $2$, and $3$.
\end{enumerate}

\textbf{Queries for Options (If the first three lockers are assigned to girls, which one of the following must be true?)}

\begin{itemize}
    \item \textbf{A) Juan is assigned to locker 1.} Query: Juan\_locker = 1 and Juan\_locker is shared with a girl
    
    \item \textbf{B) Nita is assigned to locker 3.} Query: Nita\_locker = 3
    
    \item \textbf{C) Trisha is assigned to locker 1.} Query: Trisha\_locker = 1
    
    \item \textbf{D) Juan is assigned to the same locker as Trisha.} Query: Juan\_locker = Trisha\_locker
    
    \item \textbf{E) Paul is assigned to the same locker as Trisha.} Query: Paul\_locker = Trisha\_locker
\end{itemize}

\textbf{Plan:}
To approach this constraint optimization problem regarding locker assignments for children, follow a systematic plan that accounts for all given constraints and variables. This plan will guide you through understanding the problem, applying logical reasoning, and deducing possible configurations that satisfy all conditions.

\begin{itemize}[leftmargin=*]
    \item 1. Understand the Domain and Variables: Acknowledge the domain of lockers numbered 1 to 5. Note each child's possible locker assignments (Fred, Juan, Marc, Paul, Nita, Rachel, Trisha) within this domain.
    \item 2. Comprehend Constraints: \\
    Each locker can have either one or two children assigned, but no more. \\
    Each child is assigned exactly one locker. \\
    (... More constraints ...)

    \item (... More plan ...)

    \item This plan emphasizes a logical, step-by-step approach to deducing valid locker assignments for each child, ensuring compliance with all constraints. By methodically analyzing each query in the context of these constraints, you can identify which statements about locker assignments must be true.
\end{itemize}

}

\end{tcolorbox}

\begin{tcolorbox}[breakable, title=Step-3: Solving the Problem given the context and plan,colback=gray!5,colframe=nmgray!75!black,before skip=1pt, after skip=1pt,fontupper=\linespread{0.8}\selectfont]
{\footnotesize

\textbf{Task:}

The task is a constraint optimization problem, and you need to select the most appropriate option that is true from the given options.
The domain, variables, constraints, and relevant plan will be given to you to solve the question. \\

There will be only one answer, so choose the one you think is the most likely. \\

Here is a workflow you need to follow to solve the constraint optimization problem: \\
1.Domain and Variables: Identify the possible values (domain) and variables involved. \\
2.Apply Constraints: Read and apply the specific constraints to the variables. \\
3.List Possibilities: Enumerate all potential combinations or orders that could satisfy the constraints. \\
4.Determine Order/Values: Establish the correct order or specific values for each variable as per the constraints. \\
5.Demonstrate Satisfaction of Constraints: Provide a specific example that shows all constraints are met. \\
6.Evaluate Queries: Analyze each query against the established order or values. \\
7.Identify Correct Answer: Choose the option that aligns with the determined order/values. \\
8.Present Final Answer: Clearly state the solution that satisfies the constraints and query. \\

\textbf{Domain}

\begin{itemize}
    \item Lockers: $1$ to $5$
\end{itemize}

\textbf{Variables}

\begin{itemize}
    \item Fred\_locker $\in \{1, 2, 3, 4, 5\}$
    \item Juan\_locker $\in \{1, 2, 3, 4, 5\}$
    \item Marc\_locker $\in \{1, 2, 3, 4, 5\}$
    \item Paul\_locker $\in \{1, 2, 3, 4, 5\}$
    \item Nita\_locker $\in \{1, 2, 3, 4, 5\}$
    \item Rachel\_locker $\in \{1, 2, 3, 4, 5\}$
    \item Trisha\_locker $\in \{1, 2, 3, 4, 5\}$
\end{itemize}

\textbf{Constraints}

\begin{enumerate}
    \item \textbf{Each Locker One or Two Children:} Each locker can have either one or two children assigned to it.
    
    \item \textbf{Each Child One Locker:} Each child is assigned to exactly one locker.
    
    \item \textbf{Shared Locker Boy and Girl:} If a locker is shared, it must be shared between one boy and one girl.
    
    \item \textbf{Juan Must Share:} Juan\_locker must be the same as one girl's locker.
    
    \item \textbf{Rachel Cannot Share:} Rachel\_locker must be unique and not shared with any other child.
    
    \item \textbf{Nita Not Adjacent to Trisha:} $|Nita\_locker - Trisha\_locker| \neq 1$
    
    \item \textbf{Fred Assigned to Locker 3:} Fred\_locker = 3
    
    \item \textbf{First Three Lockers to Girls:} At least one girl must be assigned to each of the lockers $1$, $2$, and $3$.
\end{enumerate}

\textbf{Queries for Options (If the first three lockers are assigned to girls, which one of the following must be true?)}

\begin{itemize}
    \item \textbf{A) Juan is assigned to locker 1.} Query: Juan\_locker = 1 and Juan\_locker is shared with a girl
    
    \item \textbf{B) Nita is assigned to locker 3.} Query: Nita\_locker = 3
    
    \item \textbf{C) Trisha is assigned to locker 1.} Query: Trisha\_locker = 1
    
    \item \textbf{D) Juan is assigned to the same locker as Trisha.} Query: Juan\_locker = Trisha\_locker
    
    \item \textbf{E) Paul is assigned to the same locker as Trisha.} Query: Paul\_locker = Trisha\_locker
\end{itemize}

\textbf{Plan:}
To approach this constraint optimization problem regarding locker assignments for children, follow a systematic plan that accounts for all given constraints and variables. This plan will guide you through understanding the problem, applying logical reasoning, and deducing possible configurations that satisfy all conditions.

\begin{itemize}[leftmargin=*]
    \item 1. Understand the Domain and Variables: Acknowledge the domain of lockers numbered 1 to 5. Note each child's possible locker assignments (Fred, Juan, Marc, Paul, Nita, Rachel, Trisha) within this domain.
    \item 2. Comprehend Constraints: \\
    Each locker can have either one or two children assigned, but no more. \\
    Each child is assigned exactly one locker. \\
    (... More constraints ...)
    \item 3. Apply all the constraints to each option.

    \item (... More plan ...)

    \item This plan emphasizes a logical, step-by-step approach to deducing valid locker assignments for each child, ensuring compliance with all constraints. By methodically analyzing each query in the context of these constraints, you can identify which statements about locker assignments must be true.
\end{itemize}

\textbf{Execution:}

Let's solve the given constraint optimization problem following the outlined plan:

\textbf{Option A: Juan is assigned to locker 1.}
\begin{itemize}
  \item \textbf{Negation of Condition for A}: Juan is not assigned to locker 1.
  \item \textbf{Test Against Constraints}: If Juan is not in locker 1, he could be in locker 2, 3, 4, or 5. Since locker 3 is taken by Fred (a boy), Juan cannot share with him due to constraint 3. If Juan shares locker 4 or 5 with a girl, this would still satisfy all constraints. Therefore, violating the condition in Option A can still lead to a valid order satisfying all the constraints. Thus, Option A does not have to be true.
\end{itemize}

\textbf{Option B: Nita is assigned to locker 3.}
\begin{itemize}
  \item \textbf{Negation of Condition for B}: Nita is not assigned to locker 3.
  \item \textbf{Test Against Constraints}: Since Fred is assigned to locker 3, Nita cannot be assigned to locker 3. This does not violate any constraints, as Nita can be assigned to any other locker that is not adjacent to Trisha's. Therefore, Option B cannot be true because it directly conflicts with constraint 7.

(... More solving steps ...)

\textbf{Conclusion:}
\item Upon evaluation, we notice that only option A must be true to satisfy all the constraint. Therefore, the final answer is A.
\end{itemize}
}
\end{tcolorbox}

\begin{tcolorbox}[breakable, title=Step-4: Verifying the translation and solving process ,colback=gray!5,colframe=nmgray!75!black,before skip=1pt, after skip=1pt,fontupper=\linespread{0.8}\selectfont]
{\footnotesize

\textbf{Task:}

Please verify the solution to the following constraint optimization problem using a detailed method. The problem consists of a set of variables, each with a specific domain, and a set of constraints. Your task is to determine if the proposed solution in the "Original execution" correctly satisfies all the constraints, and if not, find a valid solution that does.

\begin{itemize}
  \item \textbf{Context will contain the following:}
    \begin{itemize}
      \item \textbf{Domain:} Specify the domain for the problem, for example, `1: highest to 5: lowest' or `1: oldest to 5: newest'.
      \item \textbf{Variables:} List all the variables in the problem with their possible values in the domain.
      \item \textbf{Constraints:} Describe each constraint in detail, making sure to interpret them correctly according to the domain. For example, `X > Y' means `X finishes below Y' if the domain is `1: highest to 5: lowest'.
      \item \textbf{Query:} State the query that needs to be answered based on the variables and constraints.
    \end{itemize}
  \item \textbf{Your task is to apply this method to verify the problem, and if necessary, re-solve it to find a valid solution.}
  \item \textbf{Here are the requirements:}
    \begin{itemize}
      \item Please pay attention to the domain, and make sure the value for each variable is consistent with the meaning of the domain and the natural language. For instance, if the Domain: 1 (cheapest) to 7 (most expensive), and a constraint says X is the second-cheapest, then the value of X should be 2 but NOT 6 because 6 means the second-most expensive which violates the meaning of natural language. In this case, the smaller value infers the cheaper the thing is. So when you do comparison to check whether a constraint is satisfied or not, make sure you correctly interpret the direction of being cheaper is smaller value but NOT higher value. In this case, if a constraint says X is cheaper than Y. If X is 3 and Y is 2, this constraint is violated because the domain indicates that smaller value means cheaper price and 3 being larger than 2 means X is more expensive than Y, which violates the constraint.
      \item Please pay attention to the meaning of the domain, and make sure you interpret the direction of the value based on the domain correctly to particular cases.
      \item \textbf{IMPORTANT:} Review the domain, variables, constraints, and query for consistency. If there is inconsistency between the symbolic format and the natural language description of constraints, make sure you only change the symbolic format but DON'T change the natural language as natural language is the ground truth standard.
      \item Make sure you keep iterating until you find the valid order that can accurately answer the question.
      \item Make sure you fully understand the question before verification.
    \end{itemize}
\end{itemize}

\textbf{Domain}

\begin{itemize}
    \item Lockers: $1$ to $5$
\end{itemize}

\textbf{Variables}

\begin{itemize}
    \item Fred\_locker $\in \{1, 2, 3, 4, 5\}$
    \item Juan\_locker $\in \{1, 2, 3, 4, 5\}$
    \item Marc\_locker $\in \{1, 2, 3, 4, 5\}$
    \item Paul\_locker $\in \{1, 2, 3, 4, 5\}$
    \item Nita\_locker $\in \{1, 2, 3, 4, 5\}$
    \item Rachel\_locker $\in \{1, 2, 3, 4, 5\}$
    \item Trisha\_locker $\in \{1, 2, 3, 4, 5\}$
\end{itemize}

\textbf{Constraints}

\begin{enumerate}
    \item \textbf{Each Locker One or Two Children:} Each locker can have either one or two children assigned to it.
    
    \item \textbf{Each Child One Locker:} Each child is assigned to exactly one locker.
    
    \item \textbf{Shared Locker Boy and Girl:} If a locker is shared, it must be shared between one boy and one girl.
    
    \item \textbf{Juan Must Share:} Juan\_locker must be the same as one girl's locker.
    
    \item \textbf{Rachel Cannot Share:} Rachel\_locker must be unique and not shared with any other child.
    
    \item \textbf{Nita Not Adjacent to Trisha:} $|Nita\_locker - Trisha\_locker| \neq 1$
    
    \item \textbf{Fred Assigned to Locker 3:} Fred\_locker = 3
    
    \item \textbf{First Three Lockers to Girls:} At least one girl must be assigned to each of the lockers $1$, $2$, and $3$.
\end{enumerate}

\textbf{Queries for Options (If the first three lockers are assigned to girls, which one of the following must be true?)}

\begin{itemize}
    \item \textbf{A) Juan is assigned to locker 1.} Query: Juan\_locker = 1 and Juan\_locker is shared with a girl
    
    \item \textbf{B) Nita is assigned to locker 3.} Query: Nita\_locker = 3
    
    \item \textbf{C) Trisha is assigned to locker 1.} Query: Trisha\_locker = 1
    
    \item \textbf{D) Juan is assigned to the same locker as Trisha.} Query: Juan\_locker = Trisha\_locker
    
    \item \textbf{E) Paul is assigned to the same locker as Trisha.} Query: Paul\_locker = Trisha\_locker
\end{itemize}

\textbf{Original Execution:}

Let's solve the given constraint optimization problem following the outlined plan:

\textbf{Option A: Juan is assigned to locker 1.}
\begin{itemize}
  \item \textbf{Negation of Condition for A}: Juan is not assigned to locker 1.
  \item \textbf{Test Against Constraints}: If Juan is not in locker 1, he could be in locker 2, 3, 4, or 5. Since locker 3 is taken by Fred (a boy), Juan cannot share with him due to constraint 3. If Juan shares locker 4 or 5 with a girl, this would still satisfy all constraints. Therefore, violating the condition in Option A can still lead to a valid order satisfying all the constraints. Thus, Option A does not have to be true.
\end{itemize}

\textbf{Option B: Nita is assigned to locker 3.}
\begin{itemize}
  \item \textbf{Negation of Condition for B}: Nita is not assigned to locker 3.
  \item \textbf{Test Against Constraints}: Since Fred is assigned to locker 3, Nita cannot be assigned to locker 3. This does not violate any constraints, as Nita can be assigned to any other locker that is not adjacent to Trisha's. Therefore, Option B cannot be true because it directly conflicts with constraint 7.

(... More solving steps ...)

\textbf{Conclusion:}
\item Upon evaluation, we notice that only option A must be true to satisfy all the constraints. Therefore, the final answer is A.
\end{itemize}

\textbf{Verification:}

Verification of the translation and the Original Execution: \\
\textbf{Context verification:}
\begin{itemize}
    
    \item  The context and query correctly correspond to the information in the original context and are semantically consistent.

\end{itemize}

\textbf{Logical Process Verification:}
\begin{itemize}
    \item Given the constraints and the domain's interpretation, let's verify the original execution and other options:
    \item Option A: Verification of Constraints: If Juan is assigned to locker 1, he must share with a girl due to constraint 4. Since locker 3 is taken by Fred (a boy), and Rachel cannot share a locker (constraint 5), Rachel must be in locker 2. This leaves locker 1 for Juan to share with a girl other than Rachel. This satisfies all constraints, making Option A a possibility.
    \item Option B: Nita is assigned to locker 3. Verification of Constraints: Since Fred is assigned to locker 3 (constraint 7), Nita cannot be assigned to locker 3. This directly conflicts with constraint 7, making Option B impossible.
    \item (... More logical verification ...)
    \item Given that the first three lockers are assigned to girls, Fred is in locker 3, and Rachel cannot share a locker, Rachel must be in locker 1 or 2. If Rachel is in locker 1, then Trisha cannot be in locker 1, which means Trisha must be in locker 2. This leaves locker 1 available for Juan to share with a girl, which must be the case since Juan must share a locker with a girl, and locker 3 is occupied by Fred, a boy. Therefore, Juan must be assigned to locker 1 to satisfy the constraints, making Option A the one that must be true. Upon verification, we found that Option A indeed is the only possible answer. Thus, the answer A should remain unchanged.
\end{itemize}

}
\end{tcolorbox}
\subsection{Chain-of-Thought (CoT)}
Below is the prompt for CoT across each dataset.
\subsubsection{ProntoQA}
\begin{tcolorbox}[breakable,colframe=nmgray!75!black,fontupper=\linespread{0.8}\selectfont]
{\footnotesize
\noindent Given a problem statement as context, the task is to answer a logical reasoning question.

\textbf{Context:}
\begin{itemize}
    \item Jompuses are not shy.
    \item Jompuses are yumpuses.
    \item Each yumpus is aggressive.
    \item Each yumpus is a dumpus.
    \item Dumpuses are not wooden.
    \item Dumpuses are wumpuses.
    \item Wumpuses are red.
    \item Every wumpus is an impus.
    \item Each impus is opaque.
    \item Impuses are tumpuses.
    \item Numpuses are sour.
    \item Tumpuses are not sour.
    \item Tumpuses are vumpuses.
    \item Vumpuses are earthy.
    \item Every vumpus is a zumpus.
    \item Zumpuses are small.
    \item Zumpuses are rompuses.
    \item Max is a yumpus.
\end{itemize}

\textbf{Question:} Is the following statement true or false? Max is sour.

\textbf{Options:}
\begin{itemize}
    \item[A)] True
    \item[B)] False
\end{itemize}

\textbf{Reasoning:}
\begin{itemize}
    \item Max is a yumpus.
    \item Each yumpus is a dumpus. So Max is a dumpus.
    \item Dumpuses are wumpuses. So Max is a wumpus.
    \item Every wumpus is an impus. So Max is an impus.
    \item Impuses are tumpuses. So Max is a tumpus.
    \item Tumpuses are not sour. So Max is not sour.
\end{itemize}
\textbf{Conclusion:}
\begin{itemize}
    \item The correct option is: \textbf{B)}
\end{itemize}

}
\end{tcolorbox}

\subsubsection{ProofWriter}

\begin{tcolorbox}[breakable,colframe=nmgray!75!black,fontupper=\linespread{0.8}\selectfont]
{\footnotesize
\noindent Given a problem statement as contexts, the task is to answer a logical reasoning question.

\textbf{Context:}
\begin{itemize}
    \item The cow is blue.
    \item The cow is round.
    \item The cow likes the lion.
    \item The cow visits the tiger.
    \item The lion is cold.
    \item The lion is nice.
    \item The lion likes the squirrel.
    \item The squirrel is round.
    \item The squirrel sees the lion.
    \item The squirrel visits the cow.
    \item The tiger likes the cow.
    \item The tiger likes the squirrel.
    \item If something is cold then it visits the tiger.
    \item If something visits the tiger then it is nice.
    \item If something sees the tiger and it is young then it is blue.
    \item If something is nice then it sees the tiger.
    \item If something likes the squirrel and it likes the cow then it visits the tiger.
    \item If something is nice and it sees the tiger then it is young.
    \item If the cow is cold and the cow visits the lion then the lion sees the squirrel.
\end{itemize}

\textbf{Question:} Based on the above information, is the following statement true, false, or unknown? The tiger is not young.

\textbf{Options:}
\begin{itemize}
    \item[A)] True
    \item[B)] False
    \item[C)] Unknown
\end{itemize}

\textbf{Reasoning:}
\begin{itemize}
    \item The tiger likes the cow.
    \item The tiger likes the squirrel.
    \item If something likes the squirrel and it likes the cow, then it visits the tiger. So the tiger visits the tiger.
    \item If something visits the tiger then it is nice. So the tiger is nice.
    \item If something is nice and it sees the tiger then it is young. So the tiger is young.
\end{itemize}

\textbf{Conclusion:}
\begin{itemize}
    \item The correct option is: \textbf{B)}
\end{itemize}

}
\end{tcolorbox}

\subsubsection{FOLIO}
.
\begin{tcolorbox}[breakable, colframe=nmgray!75!black, before skip=0pt, after skip=0pt, fontupper=\linespread{0.8}\selectfont]
{\footnotesize
Given a problem statement as contexts, the task is to answer a logical reasoning question.

\textbf{Context:}
\begin{itemize}
    \item The Blake McFall Company Building is a commercial warehouse listed on the National Register of Historic Places.
    \item The Blake McFall Company Building was added to the National Register of Historic Places in 1990.
    \item The Emmet Building is a five-story building in Portland, Oregon.
    \item The Emmet Building was built in 1915.
    \item The Emmet Building is another name for the Blake McFall Company Building.
    \item John works at the Emmet Building.
\end{itemize}

\textbf{Question:} Based on the above information, is the following statement true, false, or uncertain? The Blake McFall Company Building is located in Portland, Oregon.

\textbf{Options:}
\begin{itemize}
    \item[A)] True
    \item[B)] False
    \item[C)] Uncertain
\end{itemize}

\textbf{Reasoning:}
\begin{itemize}
    \item The Blake McFall Company Building is another name for the Emmet Building.
    \item The Emmet Building is located in Portland, Oregon.
    \item Therefore, the Blake McFall Company Building is located in Portland, Oregon.
\end{itemize}

\textbf{Conclusion:}
\begin{itemize}
    \item The correct option is: \textbf{A)}
\end{itemize}

}
\end{tcolorbox}

\subsubsection{LogicalDeduction}
.
\begin{tcolorbox}[breakable,colframe=nmgray!75!black,before skip=0pt, after skip=0pt,fontupper=\linespread{0.8}\selectfont]
{\footnotesize
Given a problem statement as contexts, the task is to answer a logical reasoning question.

\textbf{Context:}
\begin{itemize}
    \item On a branch, there are five birds: a quail, an owl, a raven, a falcon, and a robin.
    \item The owl is the leftmost.
    \item The robin is to the left of the raven.
    \item The quail is the rightmost.
    \item The raven is the third from the left.
\end{itemize}

\textbf{Question:} Which of the following is true?

\textbf{Options:}
\begin{itemize}
    \item[A)] The quail is the rightmost.
    \item[B)] The owl is the rightmost.
    \item[C)] The raven is the rightmost.
    \item[D)] The falcon is the rightmost.
    \item[E)] The robin is the rightmost.
\end{itemize}

\textbf{Reasoning:}
\begin{itemize}
    \item The owl is the leftmost. This means the owl is not the rightmost.
    \item The robin is to the left of the raven. This means neither the robin nor the raven is the rightmost.
    \item The quail is the rightmost. This directly answers the question, so option A) The quail is the rightmost is true.
    \item The raven is the third from the left. This means the raven is also not the rightmost.
\end{itemize}

\textbf{Conclusion:}
\begin{itemize}
    \item The correct option is: \textbf{A)}
\end{itemize}

}
\end{tcolorbox}

\subsubsection{AR-LSAT}
.
\begin{tcolorbox}[breakable,colframe=nmgray!75!black,before skip=0pt, after skip=0pt,fontupper=\linespread{0.8}\selectfont]
{\footnotesize
Given a problem statement as contexts, the task is to answer a logical reasoning question.

\textbf{Context:}
\begin{itemize}
    \item During a single week, from Monday through Friday, tours will be conducted of a company's three divisions—Operations, Production, and Sales.
    \item Exactly five tours will be conducted that week, one each day.
    \item The schedule of tours for the week must conform to the following restrictions:
    \begin{itemize}
        \item Each division is toured at least once.
        \item The Operations division is not toured on Monday.
        \item The Production division is not toured on Wednesday.
        \item The Sales division is toured on two consecutive days, and on no other days.
        \item If the Operations division is toured on Thursday, then the Production division is toured on Friday.
    \end{itemize}
\end{itemize}

\textbf{Question:} Which one of the following CANNOT be true of the week's tour schedule?

\textbf{Options:}
\begin{itemize}
    \item[A)] The division that is toured on Monday is also toured on Tuesday.
    \item[B)] The division that is toured on Monday is also toured on Friday.
    \item[C)] The division that is toured on Tuesday is also toured on Thursday.
    \item[D)] The division that is toured on Wednesday is also toured on Friday.
    \item[E)] The division that is toured on Thursday is also toured on Friday.
\end{itemize}

\textbf{Reasoning:}
\begin{itemize}
    \item Since Thursday and Friday already have tours planned, only Monday, Tuesday, and Wednesday tours need to be determined.
    \item On Tuesday, the Sales division is toured since that division is toured on only two consecutive days, i.e., either Monday and Tuesday or Tuesday and Wednesday. A different division is toured on Thursday.
    \item Therefore, the final answer is C.
\end{itemize}

\textbf{Conclusion:}
\begin{itemize}
    \item The correct option is: \textbf{C)}
\end{itemize}

}
\end{tcolorbox}

\clearpage

\end{document}